\documentclass[runningheads]{llncs}

\usepackage{eccv}

\usepackage{eccvabbrv}

\usepackage{graphicx}
\usepackage{booktabs}

\usepackage[accsupp]{axessibility}  

\usepackage{hyperref}

\usepackage{orcidlink}

\usepackage{dsfont}

\newcommand{\graph}[0]{\mathcal{G}}

\newcommand{\vertexSet}[0]{\mathcal{V}}
\newcommand{\featureSet}[0]{\mathbf{X}}

\newcommand{\oneDimSpace}[1]{\mathds{R}^{#1}}

\newcommand{\loss}[0]{\mathcal{L}}

\newcommand{\gnnModel}[0]{\mathcal{M}}
\newcommand{\expect}[0]{\mathds{E}}

\usepackage{algorithm}
\usepackage{algorithmic}

\begin{document}

\title{On the Topology Awareness and Generalization Performance of Graph Neural Networks} 

\titlerunning{Topology Awareness Framework}

\author{Junwei Su
\and
Chuan Wu$^{\dagger}$
}

\authorrunning{J.Su et al.}

\institute{Department of Computer Science, the University of Hong Kong
\email{\{jwsu,cwu\}@cs.hku.hk}
}

\maketitle

\begin{abstract}
Many computer vision and machine learning problems are modelled as learning tasks on graphs, where graph neural networks (GNNs) have emerged as a dominant tool for learning representations of graph-structured data. A key feature of GNNs is their use of graph structures as input, enabling them to exploit the graphs' inherent topological properties—known as the topology awareness of GNNs. Despite the empirical successes of GNNs, the influence of topology awareness on generalization performance remains unexplored, particularly for node-level tasks that diverge from the assumption of data being independent and identically distributed (I.I.D.). The precise definition and characterization of the topology awareness of GNNs, especially concerning different topological features, are still unclear. This paper introduces a comprehensive framework to characterize the topology awareness of GNNs across any topological feature. Using this framework, we investigate the effects of topology awareness on GNN generalization performance. Contrary to the prevailing belief that enhancing the topology awareness of GNNs is always advantageous, our analysis reveals a critical insight: improving the topology awareness of GNNs may inadvertently lead to unfair generalization across structural groups, which might not be desired in some scenarios. Additionally, we conduct a case study using the intrinsic graph metric, the shortest-path distance, on various benchmark datasets. The empirical results of this case study confirm our theoretical insights. Moreover, we demonstrate the practical applicability of our framework by using it to tackle the cold start problem in graph active learning. \let\thefootnote\relax\footnotetext{$\dagger$: corresponding authors}
\let\thefootnote\relax\footnotetext{To appear as a conference paper at ECCV 2024}
\end{abstract}

\section{Introduction}\label{sec:introduction}
\begin{figure}[!t]
    \centering
  \includegraphics[width=0.98\textwidth]{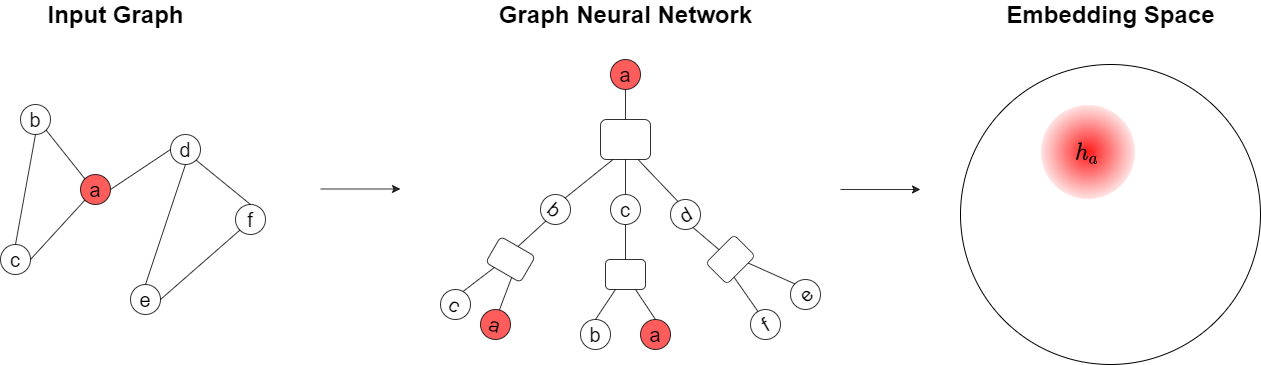}
 \caption{An illustration of the learning process in a 2-layer GNN. The message-passing mechanism leverages the graph structure to aggregate information, thereby generating the representation/embedding $h_a$ for the target vertex $a$ (highlighted in red).
 }
  \label{fig:gnn}
\end{figure}

Many problems in computer vision and machine learning are modeled as learning tasks on graphs. For example, in semantic segmentation, graphs model the relationships between different image regions, enhancing accuracy and context-aware segmentation. Graph neural networks (GNNs) have emerged as a dominant class of machine learning models specifically designed for learning representations of graph-structured data. They have demonstrated considerable success in addressing a wide range of graph-related problems in various domains such as chemistry~\cite{gilmer_quantum_chem}, biology~\cite{protein}, social networking~\cite{gnn_stochastic,sheng2024mspipe,gcn_diffusion}, scene graph generation~\cite{zhu2022scene, yang2022panoptic} and visual relationship detection~\cite{lu2016visual, xu2017scene, zellers2018neural}. A defining characteristic of GNNs is their use of a spatial approach through message passing on the graph structure for feature aggregation. This enables GNNs to preserve structural information or dependencies (referred to as {\em topology awareness}) from the underlying graph structure, allowing them to be highly effective in tasks such as node classification. Fig.~\ref{fig:gnn} illustrates the overall learning process of GNNs.

Despite their practicality and potential, there remains a lack of theoretical understanding about GNNs, particularly in the semi-supervised node classification setting where the dependencies among the data differ significantly from other machine learning models~\cite{subgroup}. In this setting, the goal is to leverage relations, as captured by the graph structure, among the data and a small set of labelled nodes to predict labels for the remaining nodes. Most of the existing theoretical studies of GNNs have focused on the connection between the message-passing mechanism of GNNs and the Weisfeiler-Lehman isomorphism test~\cite{wl_test}, aiming to understand the ability of GNNs to differentiate different graph structures in the learned representations, known as the expressive power of GNNs. Inspired by the expressiveness studies, it is commonly believed that increasing topology awareness is universally beneficial and many studies focus on enabling GNNs to preserve more structural properties in the learned representation~\cite{expressive_survey,pgnn,reach_gnn}.

 However, as GNNs become more reliant on and sensitive (aware) of graph structure as input, they may exhibit different generalization performances towards certain {\em structural subgroups} (distinct data subsets grouped by structural similarity to the training set) within the data. The quantification of GNN generalization across distinct structural subgroups is termed {\em structural subgroup generalization}~\cite{subgroup}. Such considerations are vital in GNN application and development. For instance, within protein-protein interaction networks, these structural subgroups could represent different molecular complexes, influencing the accuracy of interaction predictions. Similarly, understanding how the topology awareness of GNNs influences generalization is essential when devising sampling strategies for training. The extent to which the generalization performance of GNNs is influenced by specific structural features of graph data is critical in deciding the composition of training datasets. Despite its importance, an understanding of the relationship between the topology awareness of GNNs and its structural subgroup generalization is still lacking. Furthermore, characterizing the topology awareness of GNNs poses a challenge, especially considering that different domains and tasks may prioritize distinct structural aspects. Therefore, a versatile framework is needed to assess the topology awareness of GNNs in relation to various structures.

To address this gap, in this paper, we propose a novel framework based on approximate metric embedding to study the relationship between structural subgroup generalization and topology awareness of GNNs in the context of semi-supervised node classification.
The proposed framework allows for the investigation of the structural subgroup generalization of GNNs with respect to different structural subgroups. More concretely, the main contributions of this work are summarized as follows.
\begin{enumerate}
    \item  We propose a novel, structure-agnostic framework using approximate metric embedding to examine the interplay between GNNs' structural subgroup generalization and topology awareness. This framework is versatile, accommodating various structural measures like shortest-path distance, and requires only the corresponding structural measure. Its simplicity in estimating key factors makes it applicable and generalizable to a wide range of scenarios.
    \item Through formal analysis within our framework, we establish a clear link between GNN topology awareness and their generalization performance (Theorem~\ref{theorem:structural_relation}). We also demonstrate that while enhanced topology awareness boosts GNN expressiveness, it can result in uneven generalization performance, favouring subgroups more structurally similar to the training set (Theorem~\ref{theorem:sub_group_performance}). Such structural property can be harmful (causing unfairness issues) or useful (informing design decisions) depending on the scenario. This challenges the prevailing belief that increased topology awareness universally benefits GNNs~\cite{pgnn,reach_gnn,expressive_survey}, emphasizing the importance of considering the relationship between topology awareness and generalization performance.
    
    \item We validate our framework through a case study on shortest-path distance, highlighting its practicality and relevance. The results corroborate our theoretical findings, showing that GNNs with heightened awareness of shortest-path distances excel in classifying vertex groups closer to the training set. Moreover, we demonstrate how our findings can be applied to mitigate the cold start problem in graph active learning~\cite{hu2020graph,cold_start}, highlighting the practical implications of our framework and results. 
\end{enumerate}

\section{Related Work}\label{sec:related_work}

\subsection{Topology Awareness of GNNs} 

Modern GNN architectures take inspiration from the Weisfeiler-Lehman isomorphism test~\cite{wl_test,hamilton2020graph}, which employs the graph structure to propagate information. Consequently, a significant portion of existing literature on GNNs concentrates on understanding their ability to differentiate various graph structures, known as the expressive power of GNNs~\cite{gnn_power,higher_order_wl_gnn,pgnn,reach_gnn,group,oono2019graph}. Inspired by the
expressiveness studies, it is commonly believed that increasing topology awareness is universally beneficial and many studies focus on enabling GNNs to preserve more structural properties~\cite{su2022structure,su2023structural} in the learned representation~\cite{expressive_survey,pgnn,reach_gnn}.  Nevertheless, the precise ramifications of heightened topology awareness on GNNs' generalization performance remain shrouded in ambiguity. The intricate variations within structures across different domains and tasks further compound this complexity, underscoring the necessity for a unified framework—akin to the one proposed in this paper—to shed light on these intricacies.

\subsection{Generalization of GNNs}
Relatively few studies focus on understanding the generalization performance of GNNs in node-level tasks. \cite{subgroup,vc,su2023towards,generated_model,liao2021a,lao2022variational,yang2022graph,wu2022handling,su2024pres,esser2021learning} extend the conventional statistical learning or information-theoretic framework to GNNs, providing different generalization bounds based on concepts such as Vapnik–Chervonenkis dimension, mutual information, and algorithm stability. In light of this, despite empirical indications highlighting the critical role of GNNs' topology awareness in generalization performance\cite{pgnn,de_gnn,reach_gnn}, there remains limited understanding of the relationship between topology awareness of GNNs and their generalization. Our proposed framework addresses this gap by utilizing the distortion concept in the approximate metric embedding to connect the topology awareness of GNNs and their generalization performance. 

\subsection{Active Learning in Graph}
Active Learning (AL) seeks to enhance model performance within a limited labeling budget by identifying the most representative samples for annotation~\cite{ren2021survey}. The AL methodology generally involves two phases: 1) selecting a small, preliminary subset of samples for labeling; 2) utilizing this labeled subset to iteratively discover the most informative sample for future labeling. The initial selection of an effective subset represents the cold-start problem in AL~\cite{cold_start}. While AL has proven effective in settings with independent and identically distributed (i.i.d.) data, such as images and text, applying it to Graph Neural Networks (GNNs) poses unique challenges. These challenges stem from the necessity to account for node interdependencies and the computational complexity of iterative sample selection and retraining, especially when considering multi-hop neighbors. Recent AL methodologies for GNNs, such as AGE~\cite{cai2017active}, FeatProp~\cite{wu2019active}, and GraphPart~\cite{ma2022partition}, aim to overcome these obstacles by leveraging the graph structure and initial node features to boost efficiency and efficacy, although their success may significantly hinge on the quality of the initial features. In this work, we focus on the cold-start problem of graph active learning and introduce a novel approach (based on our analysis) that relies exclusively on graph structure, thus eliminating the reliance on feature quality at the onset of learning.

\section{Framework}\label{sec:framework}
In this section, we introduce our proposed framework, which aims to investigate the intricate relationship between topology awareness and generalization performance in GNNs. The main idea of the framework is to use metric distortion to characterize the topology awareness of GNNs. With this characterization, we are able to show the relation between topology awareness of GNNs and their generalization performance.

\subsection{GNN Preliminaries}\label{sec:preliminary}

Let $\mathcal{G} = (\mathcal{V}, \mathcal{E}, \featureSet)$ be an undirected graph where $\mathcal{V} = \{1, 2, . . . , n\}$ is the set of $n$ nodes and $\mathcal{E} \subseteq \mathcal{V} \times \mathcal{V}$ is the set of edges. Let
$\featureSet = \{x_1, x_2, . . . , x_n \}$ denote the set of node attribute vectors corresponding to all the nodes in $\vertexSet$. More specifically, $x_v \in  \featureSet$ is an m-dimensional vector which captures the attribute values of node $v \in \vertexSet$. In addition, each vertex $v \in \mathcal{V}$ is associated with a node label $y_v \in \{1,...,c\}$, where $c$ is the total number of classes.  

For our analysis, we focus on the transductive node classification setting~\cite{transductive}, where the graph $\graph$ are observed prior to learning and labels are only available for a small subset of nodes $\vertexSet_0 \subset \mathcal{V}$.  Without loss of generality, we treat $\graph$ as fixed throughout our analysis and the randomness comes from the labels $y$. Given a small set of the labelled nodes, $\vertexSet_0 \subset \vertexSet$, the task of node-level semi-supervised learning is to learn a classifier $f$, such that $\hat{y_v} = f(v)$, from a hypothesis space $\mathcal{F}$ and perform it on the remaining unlabeled nodes. Let $\loss(.,.) \mapsto \mathds{R}_+$ be a bounded continuous loss function. We denote by $R^{\loss}(f,\vertexSet)$ the generalization risk of the classifier $f$ with respect to $\loss$ and $\vertexSet$ and it is defined as follows,
\begin{equation}\label{eq:generalization_risk}
    R^{\loss}(f,\vertexSet) = \expect_{v \sim \vertexSet}[\loss(f(v), y_v)]
\end{equation}

In our case, the hypothesis space is captured by a GNN and an additional read-out function. GNNs are commonly used as an embedding function and adopt a neighbourhood aggregation scheme, where the representation of a node is learnt iteratively by aggregating representations of its neighbours. After k iterations of aggregation, a node's representation captures the information within its k-hop neighbourhood in the graph. Formally, the k-th layer of a GNN can be written as follows,
\begin{equation}
    h_{u}^{(k)} = \mathrm{Update}^{(k)} \big ( \mathrm{Aggregate}^{(k)}(\{h_v^{(k-1)}, \forall v \in \mathcal{N}(u) \cup \{u\} \}) \big ),
\end{equation}
where $h_u^{(k)}$ is the representation vector of node u at the k-th layer. $h_u^{(0)}$ is initialized to be $x_u$, and $\mathcal{N}(u)$ is the set of nodes directly adjacent to $u$. Different GNNs are compositions of aggregation and update layers of different designs. A GNN is commonly interpreted as an embedding function that combines the graph structure and node features to learn a representation vector $h_v$ of node $v$. Then the learnt representation is fed into a task-specific read-out function $g$ (commonly a multilayer perceptron) for the prediction. $g$ maps the embedding vector $h_v$ to a label $y_v$ in the label set $\oneDimSpace{c}$. The overall GNN prediction process can be viewed as a composition of $g \circ \gnnModel: \vertexSet \mapsto \oneDimSpace{c}$.  The overall GNN prediction process can be viewed as a composition of $g \circ \gnnModel: \vertexSet \mapsto \oneDimSpace{c}$.

\subsection{Structural Subgroup}
To study the generalization performance of GNNs on different structural subgroups, we assume the existence of a generalized distance (structural measure) function, $d_s(u,v)$, that captures the similarity between two nodes $u,v$ with respect to structure $s$. Furthermore, we assume that this structural measure function $d_s(u,v)$ and the vertex set $\vertexSet$ form a metric space. A metric space is a fundamental mathematical object that allows for studying and measuring the relation among the elements in a given set, and it is formally defined as follows.
\begin{definition}[metric space]\label{def:metric_space}
    A metric space is a pair $(\mathcal{Q},d)$, where $\mathcal{Q}$ is a set and $d:\mathcal{Q} \times \mathcal{Q} \mapsto \mathds{R}$ is a metric satisfying the following axioms ($x,y,z$ are arbitrary points of $\mathcal{Q}$):
        \begin{align*}
        \centering
          \text{(M1)} & d(x,y) \geq 0,   &\text{(M2)}&d(x,x) = 0,  & \text{(M3)} d(x,y) > 0,  x \neq y,\\
           \text{(M4)} & d(x,y) = d(y,x),  & \text{(M5)}&d(x,y)+d(y,z) \geq d(x,z). & 
        \end{align*}
\end{definition}

These properties ensure that the metric $d(.)$ can be used to quantify the distance or dissimilarity between any two elements in the set $\mathcal{Q}$. In the context of graph learning, the metric space $(\mathcal{V}, d_s)$ allows us to study the similarity between nodes in terms of the structural property $s$ and measure the structural distance between different subgroups of the test set. One natural and commonly used example in the graph is the metric space formed by the vertex set and the shortest-path distance.

\begin{definition}[Structural Group Distance]\label{def:distance}
    Given a metric space $(\vertexSet,d_s)$ with respect to structure $s$ and two distinct sets (groups) $\vertexSet_1, \vertexSet_2 \subset \vertexSet$, we define the structural distance $D_s$ between an element $v \in \vertexSet$ to a group as    
    \begin{equation}
        D_s(v,\vertexSet_1) =  \min_{u \in \vertexSet_1} d_s(v,u)
    \end{equation}
    and the structural distance between the two groups as
    \begin{equation}\label{eq:set_dist}
        D_s(\vertexSet_1,\vertexSet_2) = \max_{v \in \vertexSet_1}\min_{u \in \vertexSet_2} d_s(v,u)
    \end{equation}
\end{definition}

It should be noted that the distance function in Def.~\ref{def:distance} is not symmetrical, and Eq.~\ref{eq:set_dist} captures the maximal structural difference of vertices in $\vertexSet_1$ to $\vertexSet_2$. To provide further insight into the concept of a structural subgroup, let's consider a scenario where $\vertexSet_0 \subset \vertexSet$ represents the training set and $d_s(.)$ represents the shortest-path distance. In this context, we can partition the remaining vertices $\vertexSet \backslash \vertexSet_0$ into different structural subgroups based on their shortest-path distance to the nearest vertex in the training set. Specifically, we can define $\vertexSet_k = \{v \in \vertexSet \backslash \vertexSet_0 | D_s(v,\vertexSet_0) = k \}$. However, it is important to emphasize that our analysis does not assume a specific partition scheme, and the framework remains adaptable to various scenarios.

\subsection{Topology Awareness}
To characterize and formalize the topology awareness of GNNs, we introduce the concept of distortion, which measures the degree to which the graph structure is preserved in the embedding space. Mathematically, the distortion of a function between two general metric spaces is defined as,

\begin{definition}[distortion]\label{def:distortion}
    Given two metric spaces $(\mathcal{Q},d)$ and $(\mathcal{Q'},d')$ and  
 	a mapping $\gamma: \mathcal{Q} \mapsto \mathcal{Q'}$ ($\mathcal{Q}$ ($\mathcal{Q}'$) is a set and $d$ ($d'$) is a distance function)), $\gamma$  is said to have a distortion $\alpha\ge 1$, 
 	if there exists a constant $r > 0$ such that $\forall u,v \in \mathcal{Q}$, 
  $$r d(u,v) \leq d'(\gamma(u),\gamma(v))\leq \alpha r d(u,v).$$
\end{definition}



The definition above captures the maximum expansion and minimum contraction between the two metric spaces. In the extreme case when $\alpha = 1$, the mapping $\gamma$ is the so-called isometric mapping that perfectly preserves the distance between any two points in the space. In other words, for any two points $x$ and $y$ in the domain of the function, their distance $d(x, y)$ in the original metric space is equal to the distance $d'(\gamma(x), \gamma(y))$ in the target metric space (up to a constant scaling factor). If the distortion rate of the mapping is close to 1, then this mapping is said to have a low distortion rate. 

In the context of GNNs, evaluating distortion involves a straightforward comparison of embeddings in the embedding domain with the structural measures on the graph, as demonstrated in the case study.  If the distortion rate between the graph domain and the embedding domain is close to 1, the resulting embedding would largely preserve the corresponding structural information from the graph space, thereby indicating high topology awareness. Consequently, distortion can be employed to measure the topology awareness of GNNs concerning the corresponding structural properties (e.g., shortest-path distance) measured in the graph domain. This concept of distortion is agnostic to specific structural configurations, enabling the study of diverse graph structures using appropriate distance functions. Moreover, its estimation can be readily achieved by comparing the embedding with the original structural measure, as illustrated later in the case study.


\section{Main Results}\label{sec:main_result}
In this section, we present the main results derived from our proposed framework for studying the generalization and fairness of GNNs with respect to different structural groups in a graph. We start by demonstrating the relationship between the generalization performance of the model and the structural distance between the subgroups and the training set (Theorem~\ref{theorem:structural_relation}). We then show that such a structural relation can induce disparities in the generalization performance across different subgroups (Theorem~\ref{theorem:sub_group_performance}). Finally, we connect these theoretical results to the recent deep learning spline line theory~\cite{spline_theory} to provide additional insights. Detailed proof of the results presented in this paper can be found in the supplementary material. 


\subsection{Generalization Performance and Structural Distance}
\begin{theorem}\label{theorem:structural_relation}
    Let $(\vertexSet, d_s)$ be a metric space for structure $s$. Let $\vertexSet_0$ be a given labelled training set and $\vertexSet_i \subset \vertexSet \backslash \vertexSet_0$ be an arbitrary test subgroup. Suppose $\gnnModel$ is a GNN model with distortion $\alpha$ with respect to the structural measure $d_s(.,.)$, and $g$ and $\loss$ are Lipschitz,  we have that 
    \begin{equation}
        R^{\loss}(g \circ \gnnModel,\vertexSet_i) \leq R^{\loss}(g \circ \gnnModel,\vertexSet_0) + \mathcal{O} (\alpha  D_s(\vertexSet_i,\vertexSet_0)).
    \end{equation}
\end{theorem}


Theorem~\ref{theorem:structural_relation} states that the generalization risk of a classifier with respect to a structural subgroup depends on the empirical training loss of the classifier, the structural distance between the subgroup and the training set, and the distortion (topology awareness) of GNNs. Such a relation provides a new perspective on understanding the link between topology awareness and generalization performance in GNNs. Prior research in this area has mainly focused on the connection between GNNs and the Weisfeiler-Lehman algorithm, emphasizing the ability of GNNs to differentiate various graph structures. However, Theorem~\ref{theorem:structural_relation} highlights a direct link between topology awareness and generalization performance, and a potential issue of fairness in GNNs, as it shows that the model's performance on different structural subgroups is influenced by their distance to the training set. Next, we extend the result of Theorem~\ref{theorem:structural_relation} to analyze the relation between the generalization of different structural subgroups and their distance to the training set.

\subsection{Unfair Generalization Performance on Structural Subgroups}
\begin{theorem}\label{theorem:sub_group_performance}
      Let $(\vertexSet, d_s)$ be a metric space for structure $s$. Suppose $\gnnModel$ is a GNN model with small training loss on $\vertexSet_0$ and distortion $\alpha$ with respect to structural measure $d_s(.,.)$,and $\loss$ and $g$ are smooth.
    Let $\vertexSet_i, \vertexSet_j \subset \vertexSet \backslash \vertexSet_0$ be two distinct subgroups in the local neighbourhood of the training set $\vertexSet_0$, then there exists constant $\delta > 0$ (independent of the choice of subgroups) such that if 
    $$D_s(\vertexSet_i, \vertexSet_0) \geq \alpha \delta  D_s(\vertexSet_j, \vertexSet_0), $$  
    we have that, 
	$$R^{\loss}(g \circ \gnnModel,\vertexSet_i) \geq R^{\loss}(g \circ \gnnModel,\vertexSet_j).$$
\end{theorem}

Theorem \ref{theorem:sub_group_performance} states that when comparing two subgroups, if the difference in their structural distance to the training set is larger than the threshold ($\delta \alpha$), then GNNs exhibit unfair generalization performance with respect to these groups. Specifically, the theorem states that GNNs generalize better on subgroups that are structurally closer to the training set, which leads to accuracy disparities among different subgroups. Additionally, the theorem notes that the topology awareness of GNNs (measured by the distortion rate $\alpha$) plays a key role in determining the strength of this accuracy disparity. This has important implications for the use of GNNs in critical applications, as it suggests that increased topology awareness in GNNs can lead to improved generalization performance but also a greater potential for unfairness in terms of accuracy disparities among different structural subgroups. As such, it is important to carefully consider the trade-offs between topology awareness and fairness when designing new GNNs and applying them to real-world problems.

\begin{wrapfigure}{R}{0.5\textwidth}
    \centering
    \includegraphics[width=0.48\textwidth]{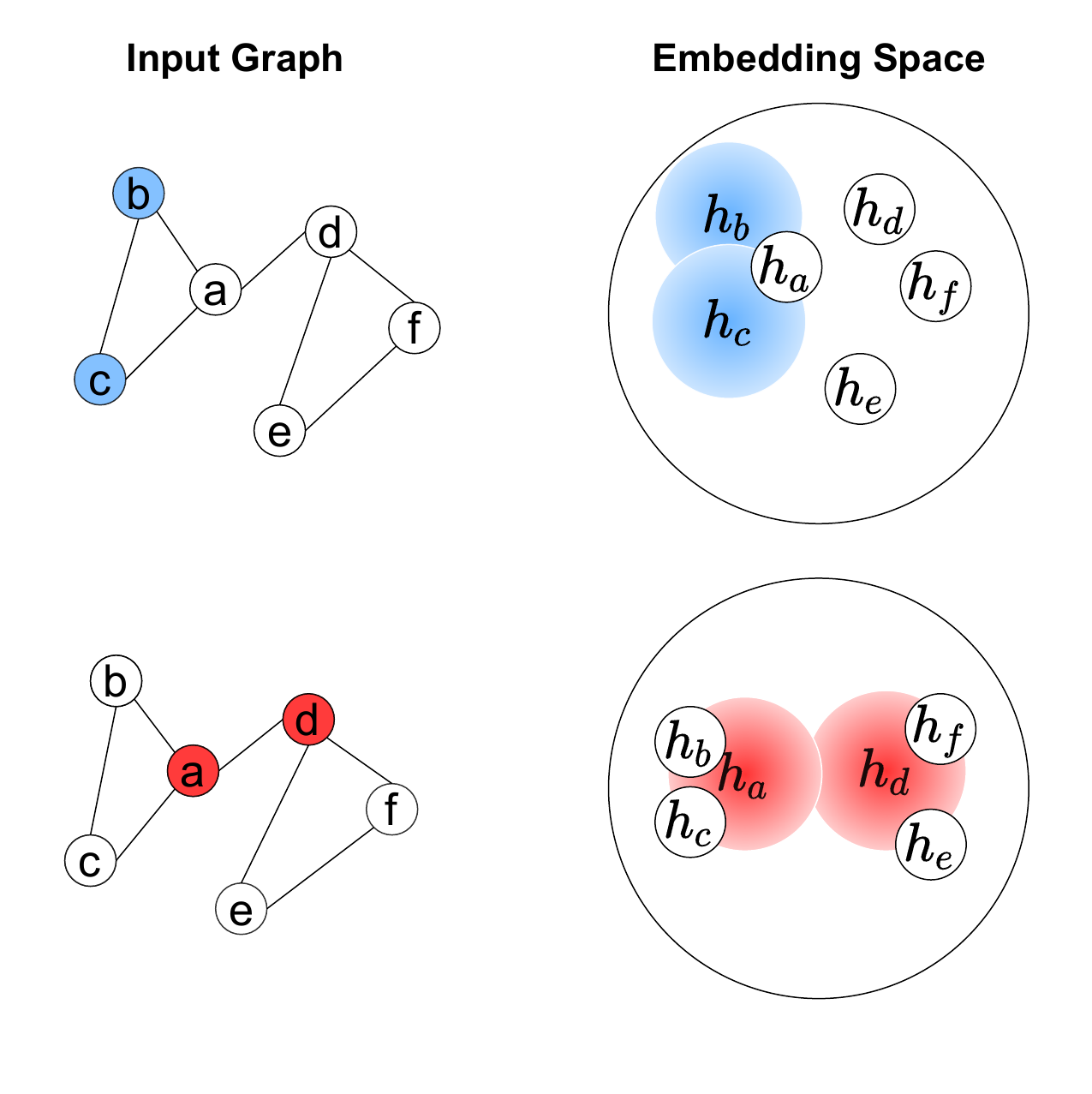}
  \caption{The loss landscape in embedding space induced by different training sets. When vertices $a$ and $d$ are selected as the training set instead of $b$ and $c$, the test losses are smaller as the average distance between the training set and the other vertices is smaller (the neighbourhoods of $h_a$ and $h_d$ cover embeddings of more vertices in the embedding space).}
  \label{fig:different_train_set}
  \vspace{-3mm}
\end{wrapfigure}


\subsection{Connection with Deep Learning Spline Theory}
Our theoretical findings reveal that when considering the performance of the model from the viewpoint of the embedding space, the training set creates a ``valley'' effect on generalization performance, with the training vertex at the bottom of the valley. Representations that are closer to the embedding of the training set perform better, while those farther away perform worse. This pattern has an interesting connection with deep learning spline theory \cite{spline_theory}. This theory suggests that in DNNs, the learning process for classification divides the input/latent space into small circles (referred to as power diagrams), where representations within each circle are grouped, and representations closer to the centre of the circle are predicted with more confidence. Our findings suggest that in GNNs, the mapping of test data to these power diagrams is related to the structural similarity between test and training data in the graph. This connection provides further validation for our derived results.


\section{A Case Study on Shortest-Path Distance}\label{sec:experiment}
In this section, we present a case study which utilizes the shortest path distance to validate the predictions and effectiveness of our proposed framework. We aim to demonstrate how our framework can be applied to a real-world scenario and how it can be used to understand the generalization performance of GNNs on different structural subgroups. The case study will provide insights and show the potential impact of our results on various critical applications of GNNs. In this case study, we aim to answer empirically the following questions: Given a GNN model that can preserve certain structure $s$ ($\alpha$ closed to 1), is the derived structural result observed on the generalization performance of the GNN model with respect to structural subgroups defined by $s$? Furthermore, how can our outcomes be harnessed to effectively tackle real-world challenges?

Overall, our case study demonstrates that our proposed framework can be applied to a real-world scenario and provides useful insights into the performance of GNNs in different structural subgroups. It also shows that the results of our framework can be generalized for different train/test pairs. Moreover, we showcase its practical utility by employing the findings to address the cold start problem within graph active learning.

{\bf General Setup.}  We use Deep Graph Library~\cite{dgl} for our implementation and experiment with four representative GNN models, GCN~\cite{gcn}, Sage~\cite{sage}, GAT~\cite{gat} and GCNII~\cite{gcn2}. In all experiments, we follow closely the common GNN experimental setup, and multiple independent trials are conducted for each experiment to obtain average results. We adopt five popular and publicly available node classification datasets: Cora~\cite{common_data}, Citeseer~\cite{common_data2}, CoraFull~\cite{cora_full}, CoauthorCS~\cite{coau_cs} and Ogbn-arxiv~\cite{ogb}. Due to space limitations, we only show part of the experimental results in the main body of the paper. Detailed descriptions of the datasets, hyperparameters and remaining experimental results can be found in the appendix.

{\bf Structural Subgroups.} We examine the accuracy disparity of subgroups defined by the shortest path distance (spd). In order to directly investigate the effect of structural distance on the generalization bound (Theorem~\ref{theorem:structural_relation}), we first split the test nodes into subgroups by their shortest-path distance to the training set. For each test node $v$, its distance to the training set $\vertexSet_0$ is given by $d_v = D_{\mathrm{spd}}(v,\vertexSet_0)$ with the shortest path distance metric as defined in Def.~\ref{def:distance}. Then we group the test nodes according to this distance. 

\begin{figure*}[!h]
\centering
\subfigure[GCN, Cora]{
\includegraphics[width=.225\textwidth]{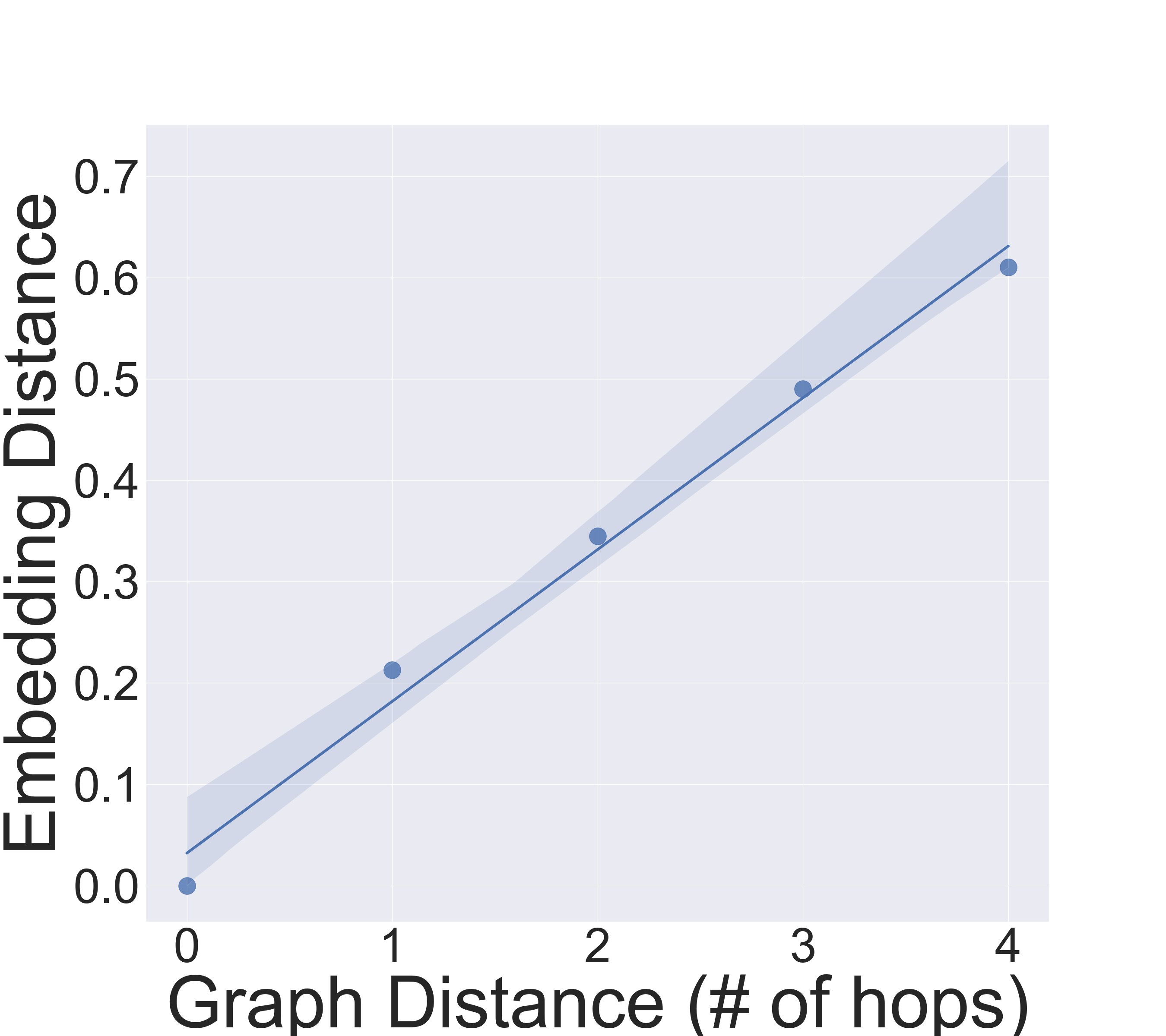}
}
\subfigure[Sage, Cora]{
\includegraphics[width=.225\textwidth]{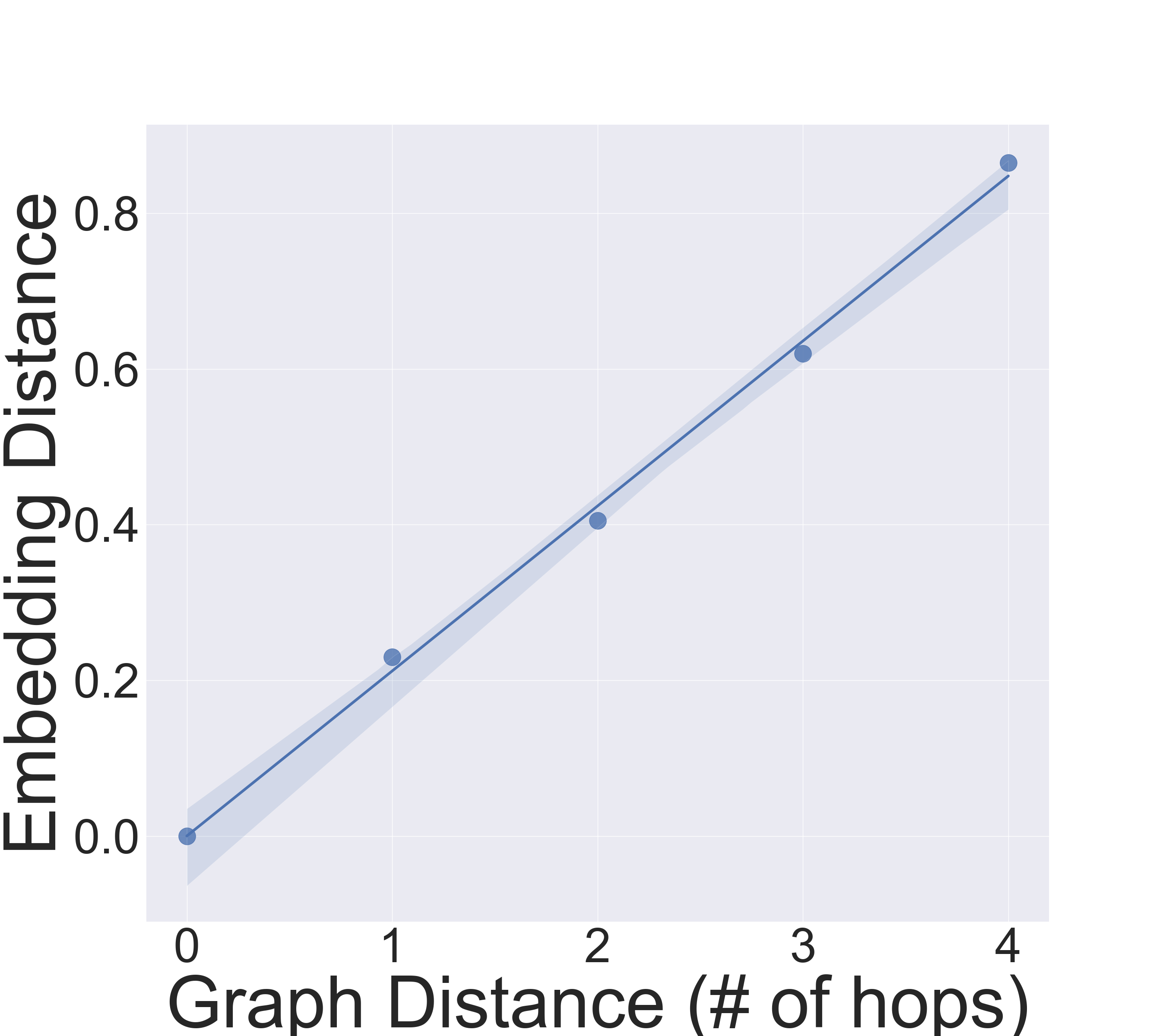}
}
\subfigure[GAT, Cora]{
\includegraphics[width=.225\textwidth]{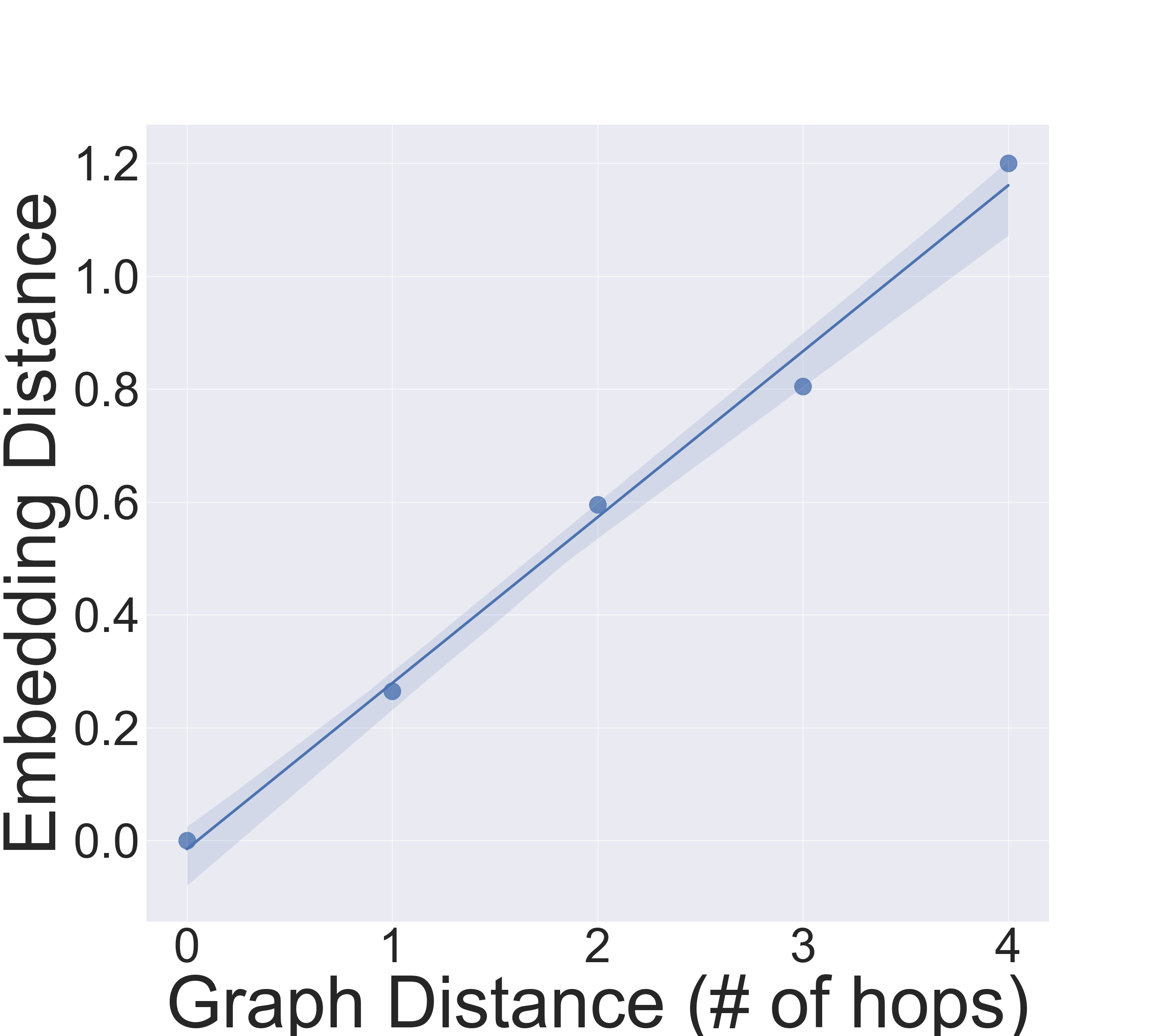}
}
\subfigure[GCN2, Cora]{
\includegraphics[width=.225\textwidth]{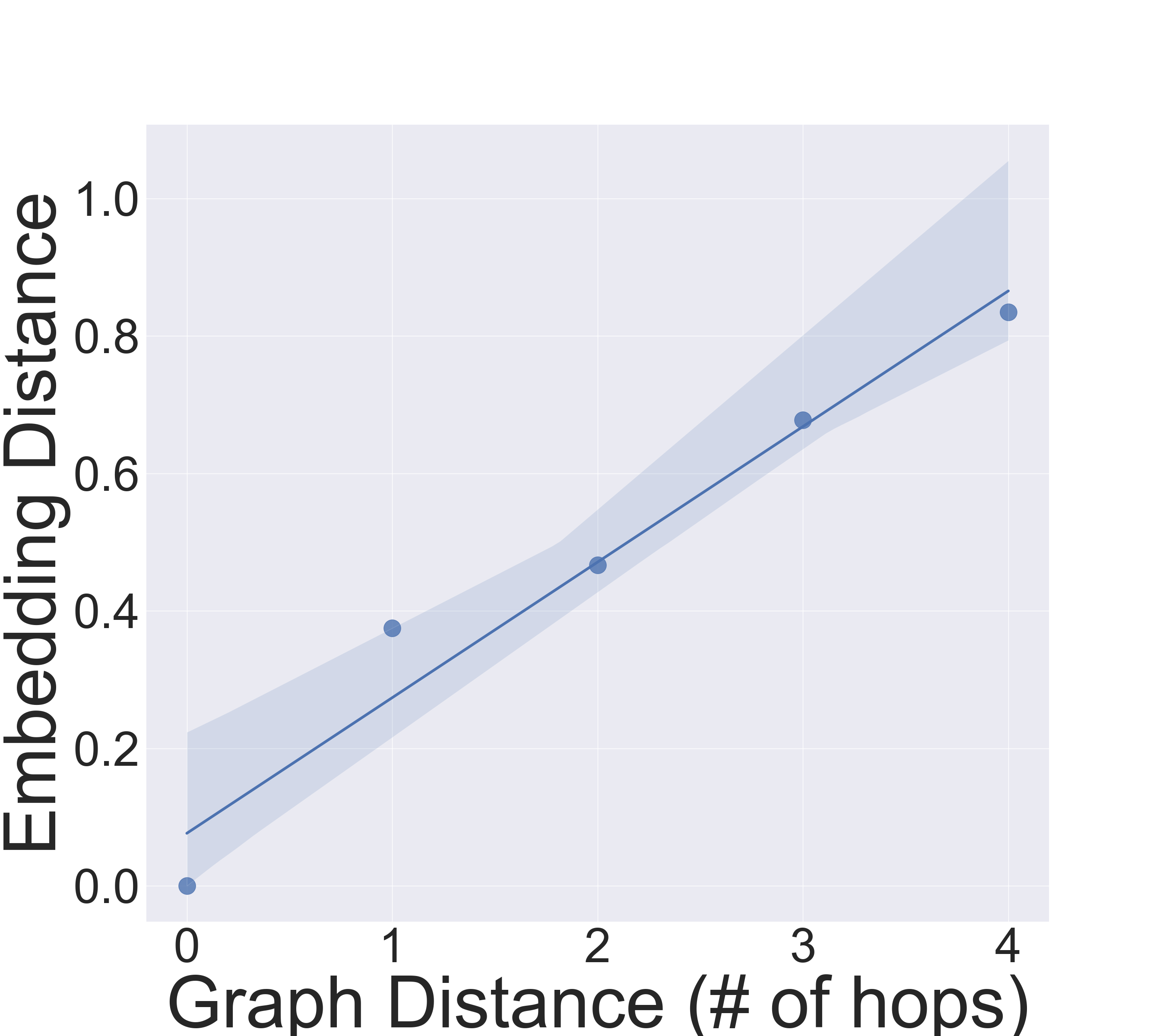}
}

\subfigure[GCN, Coauthor CS]{
\includegraphics[width=.225\textwidth]{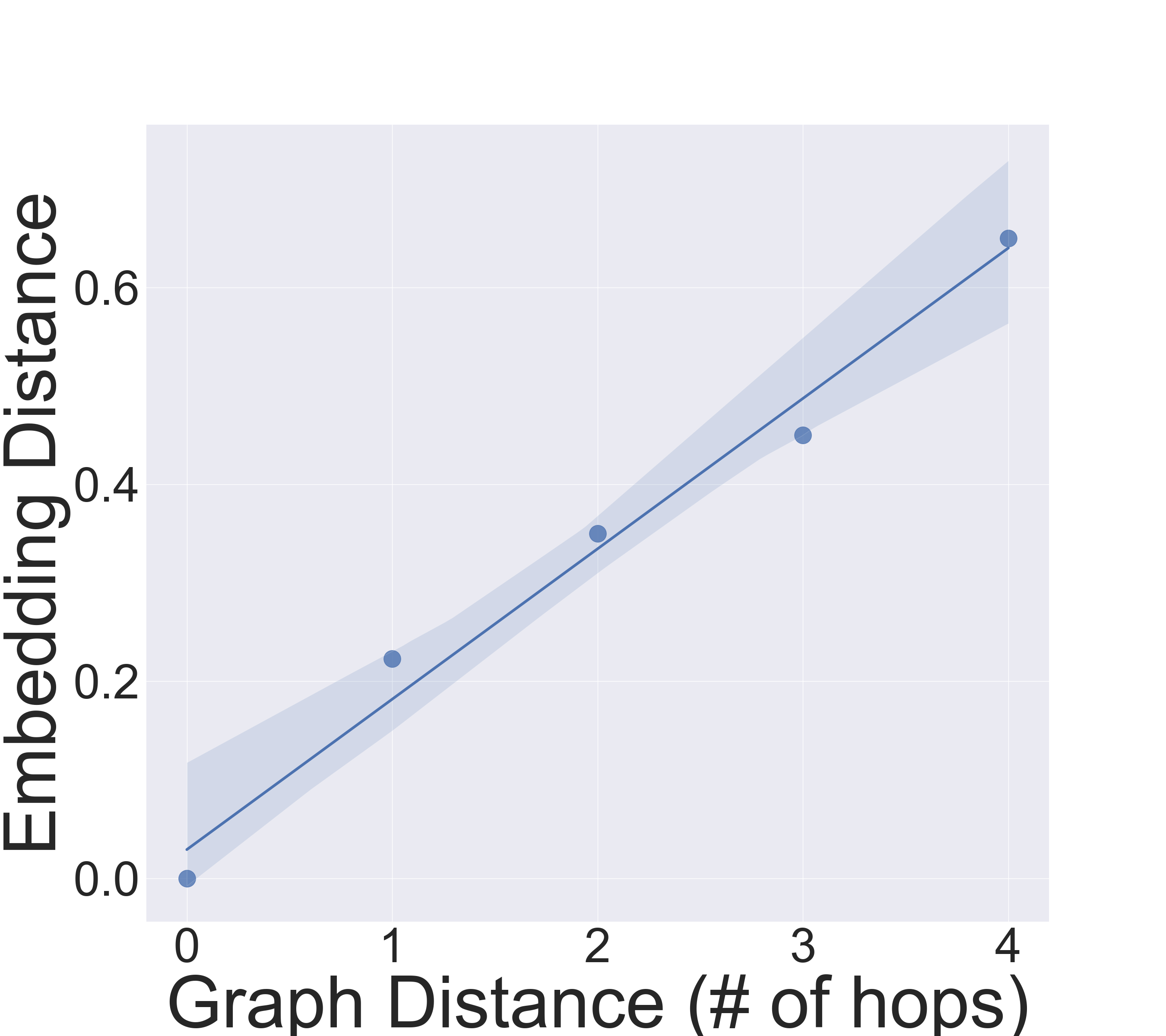}
}
\subfigure[Sage, Coauthor CS]{
\includegraphics[width=.225\textwidth]{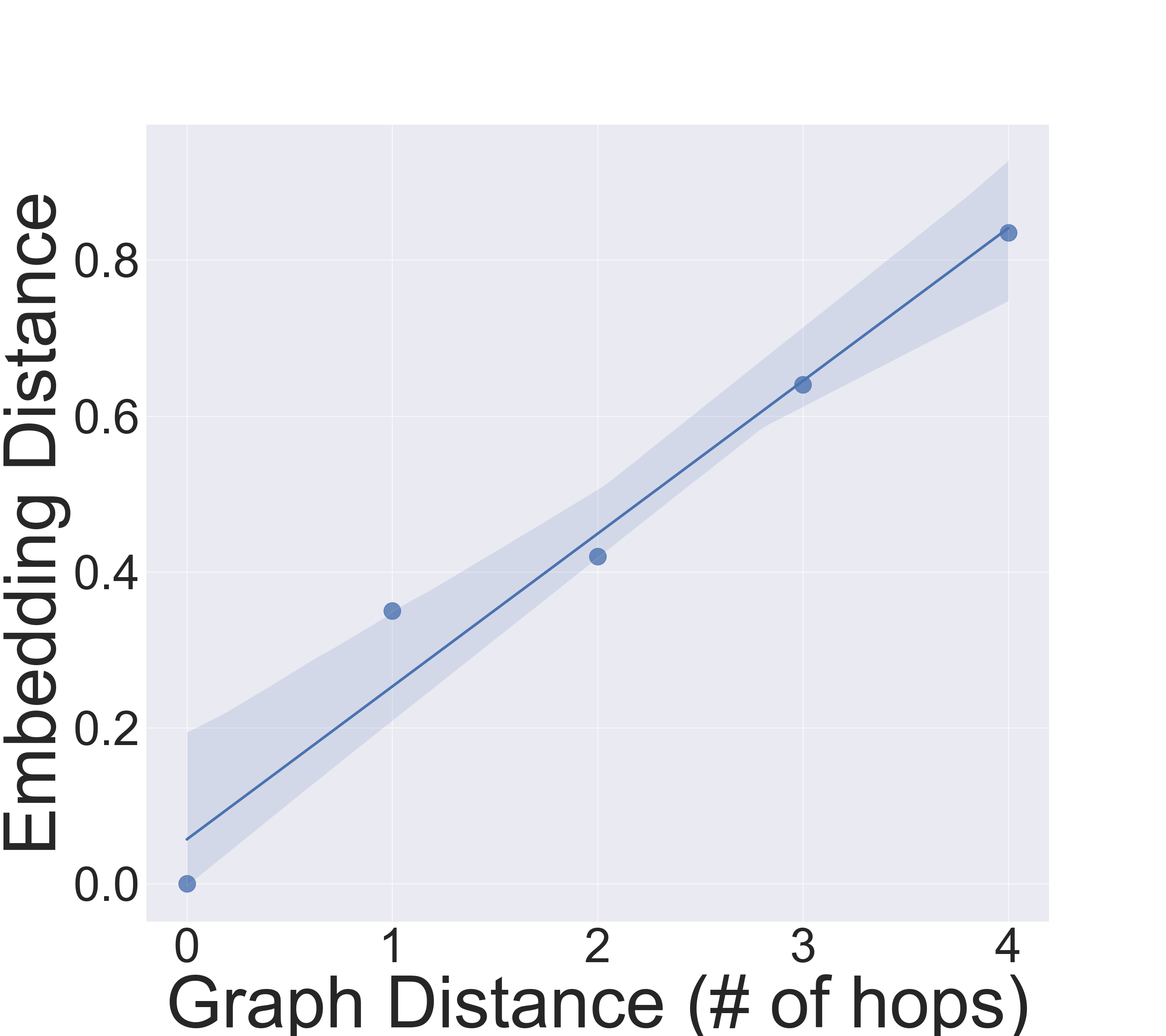}
}
\subfigure[GAT, Coauthor CS]{
\includegraphics[width=.225\textwidth]{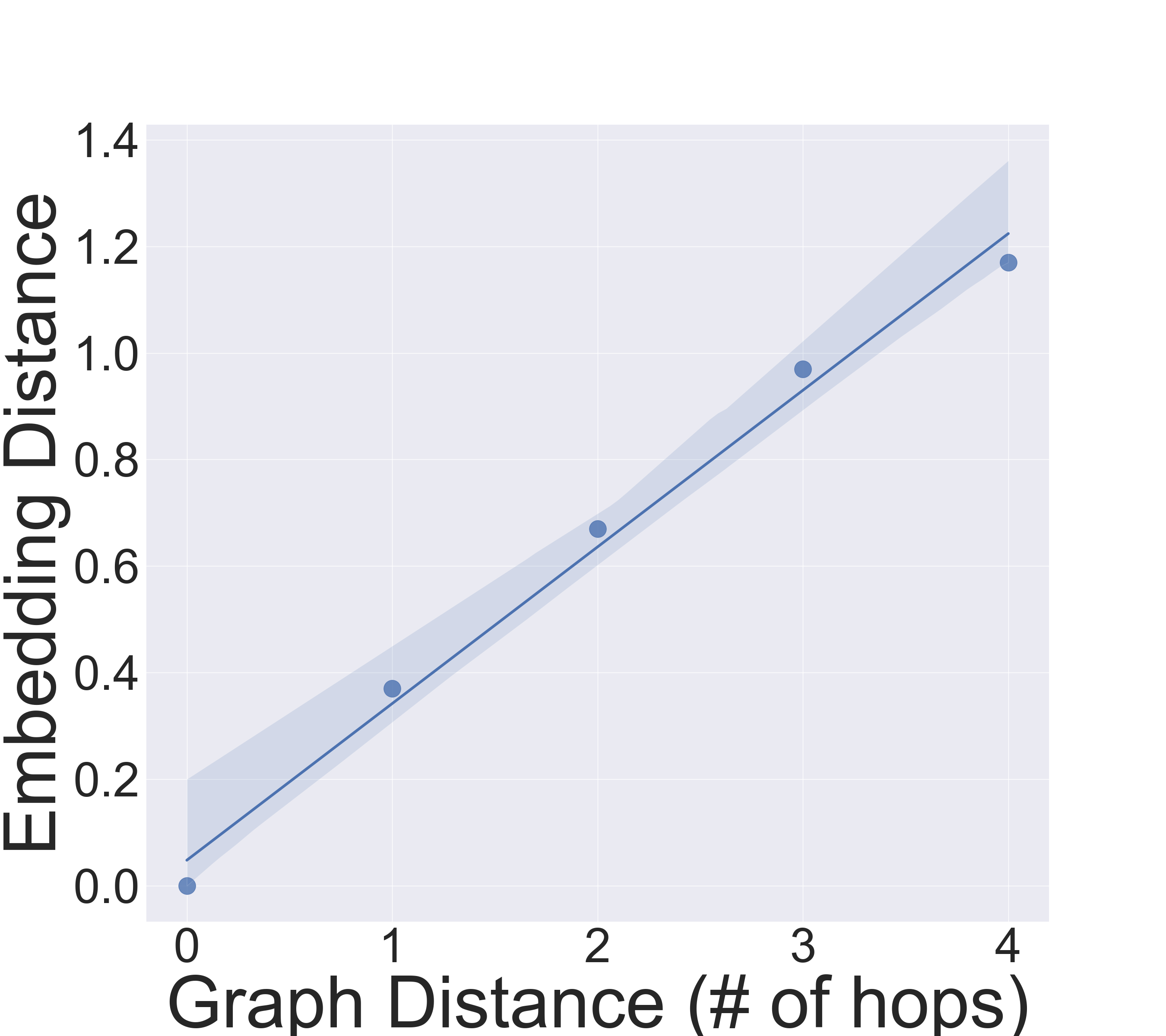}
}
\subfigure[GCNII, Coauthor CS]{
\includegraphics[width=.225\textwidth]{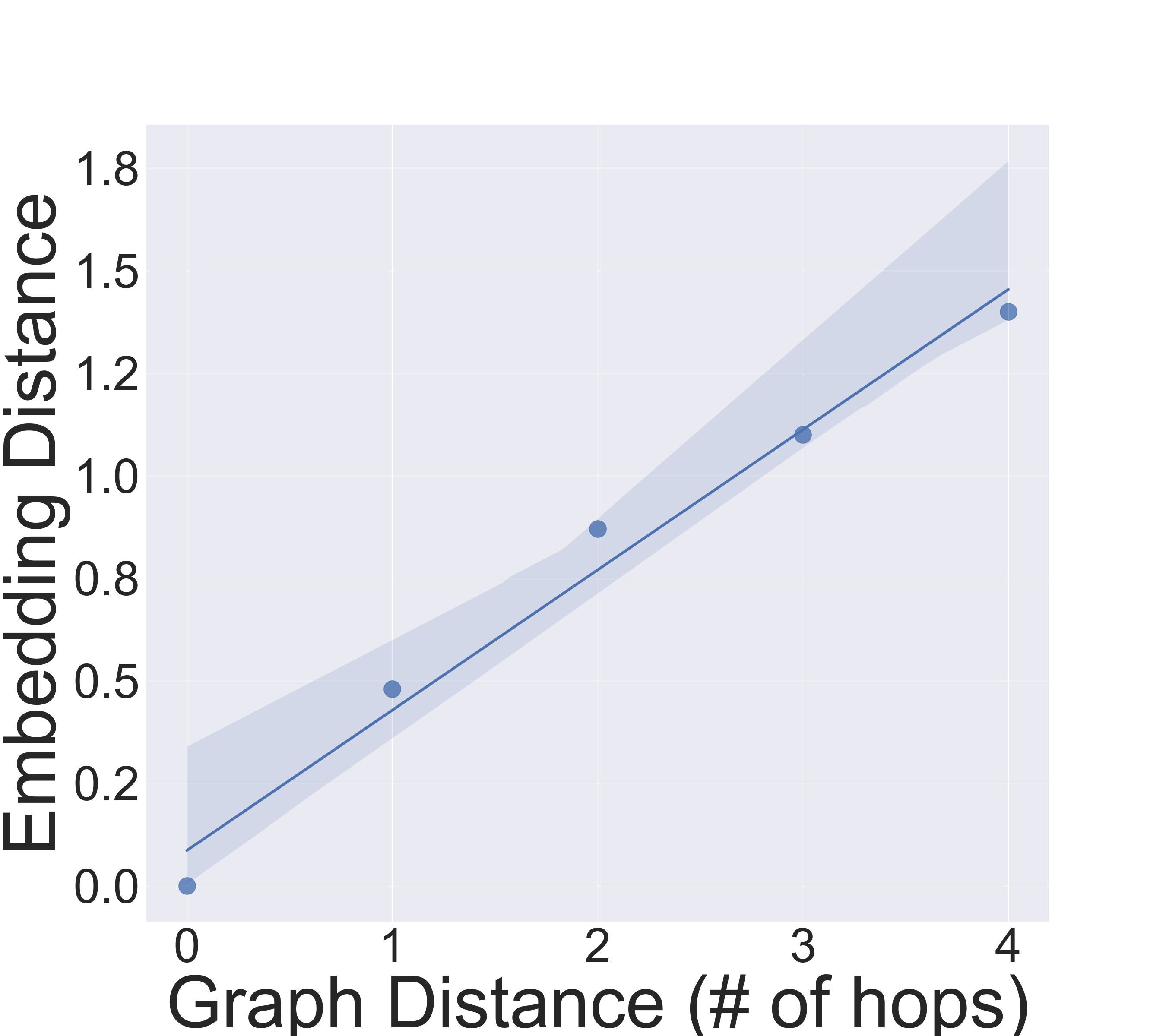}
}
\caption{Graph distance vs.~embedding distance.  (a)-(d) are the relation between graph distance (shortest-path distance) and embedding distance (Euclidean) of the four representative GNNs on Cora. (e)-(h) are the relation between graph distance and embedding distance of the four representative GNNs on Coauthor CS. The highlighted regions represent the 95\% confidence interval for that regression. (a)-(h) all demonstrate a rather linear relation between graph distance and embedding distance. }
\label{fig:graph_embed_dist}
\end{figure*}

 {\bf Distance Awareness of the Selected GNNs. }  We first verify if the selected GNNs preserve (are aware of) graph distances (with a small distortion rate), by evaluating the relation between the graph distances of test vertices to the training set and the embedding distances (Euclidean distance) of vertex representations to the representations of the training set, i.e., the relation between $d_{\mathrm{spd}}(v,\vertexSet_0)$ and $d_{\mathrm{Euclidean}}(\gnnModel(v),\gnnModel(\vertexSet_0))$. In this experiment, we train the GNN models on the default training set provided in the dataset. We extract vertices within the 5-hop neighbourhood of the training set and group the vertices based on their distances to the training set. 5-hop neighbourhood is chosen because the diameter of most real-life networks is around 5 and it gives a better visual presentation. We then compute the average embedding distance of each group to the training set in the embedding space. Based on Def.~\ref{def:distortion}, if an embedding has low distortion, we should expect to see linear relation with the slope being the scaling factor. As shown in Fig.~\ref{fig:graph_embed_dist}, there exists a strong correlation between the graph distance and the embedding distance. The relation is near linear and the relative orders of distances are mostly preserved within the first few hops. This implies that the distortion $\alpha$ is small with the selected GNNs, i.e., they are distance aware. It also indicates that the premise of the small distortion in our theoretical results can be commonly satisfied in practice. 

\subsection{Graph Distance and Generalization Performance}
\begin{figure*}[!h]
\centering
\subfigure[Cora]{
\includegraphics[width=.45\textwidth]{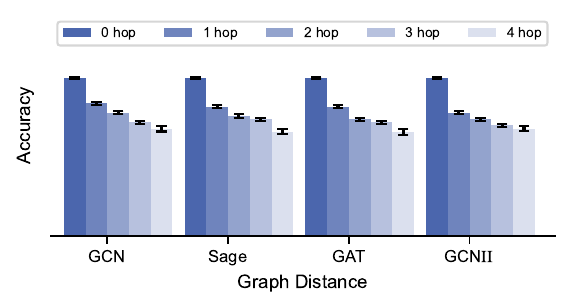}
}
\subfigure[Coauthor CS]{
\includegraphics[width=.45\textwidth]{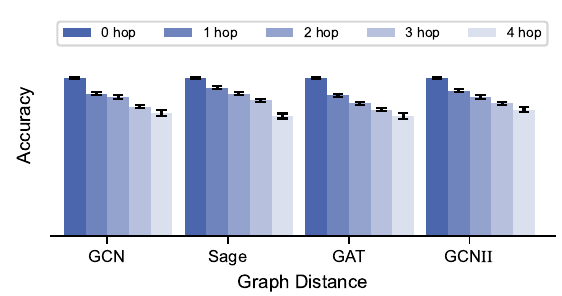}
}
\caption{Graph distance vs.~accuracy. (a) is the relation between graph distance (shortest-path distance) and the accuracy of four representative GNNs on Cora. (b) is the relation between graph distance and the accuracy of four representative GNNs on Coauthor CS. Both (a) and (b) show that the graph distance is indeed a good indicator for generalization performance when GNNs are distance aware.}
\label{fig:group_accuracy}
\end{figure*}

Next, we study the relation between graph distance and generalization performance following the same settings as in the previous experiment. We compute the average prediction accuracy in each group of vertices. In Fig.~\ref{fig:group_accuracy}, `0 hop' indicates the training set. We observe that the selected GNNs (with low distortion) indeed tend to generalize better on the group of vertices that are closer to the training set, which validates Theorem~\ref{theorem:sub_group_performance}. 
 
We further validate if the distance between different pairs of the training sets and the test group is related to the generalization performance of GNNs in the test group. We randomly sample $k=5$ vertices from each class as the training set $\vertexSet_0$. Then we evaluate the GNN model on the rest of the vertices $\vertexSet \backslash \vertexSet_0$. We make $k$ relatively small to enable a larger variance (among different trials) in the distance $D_{\mathrm{spd}}(\vertexSet \backslash \vertexSet_0 
 ,\vertexSet_0)$ (as given in Def.~\ref{def:distance}) between the training set and the rest of the vertices. We conduct 30 independent trials each selecting a different training set and ranking all the trials in descending order of the mean graph distance between the training set and the rest of the vertices. In Table~\ref{tab:mean_graph}, Group 1 contains the 10 trials with the largest graph distance $D_{\mathrm{spd}}(\vertexSet \backslash \vertexSet_0 
 ,\vertexSet_0)$ and Group 3 includes the 10 trials with the smallest graph distance $D_{\mathrm{spd}}(\vertexSet \backslash \vertexSet_0 
 ,\vertexSet_0)$. The average classification accuracy and the variance (value in the bracket) are computed for each group. 

As shown in Table~\ref{tab:mean_graph}, training GNNs on a training set with a smaller graph distance $D_{\mathrm{spd}}(\vertexSet \backslash \vertexSet_0 
 ,\vertexSet_0)$ to the rest of the vertices, does generalize better, which validates our result and further enables us to use these insights for designing and developing new GNNs algorithms.

\begin{table*}[!h]
  \centering
  \caption{Accuracy of 30 independent trials with different train/test pairs grouped by $D_{s}(\vertexSet \backslash \vertexSet_0 
 ,\vertexSet_0)$.  Group 1 contains the 10 trials with the largest graph distance $D_{\mathrm{spd}}(\vertexSet \backslash \vertexSet_0 ,\vertexSet_0)$ and Group 3 includes the 10 trials with the smallest graph distance $D_{\mathrm{spd}}(\vertexSet \backslash \vertexSet_0 ,\vertexSet_0)$. The average accuracy and variance of the 10 trials are recorded in the table. The results show that train/test pairs with closer structural similarity indeed have better generalization performance, as predicted by our theoretical results.
  }  
\resizebox{\textwidth}{!}
  {
   \setlength\tabcolsep{1.5pt}
    \begin{tabular}{c|cccc|cccc}
    \hline
          &        \multicolumn{4}{c}{Cora}              & \multicolumn{4}{c}{Citeseer}            \\
            
    \hline
          &   GCN & GAT & Sage & GCNII & GCN & GAT & Sage & GCNII\\
         
    \hline
     Group 1  &  65.42 (5.03) & 68.03 (4.62) & 67.61 (5.02) &  61.42 (4.03) & 59.72 (5.41) & 56.78 (4.73) & 58.52 (4.85) & 55.46 (3.69) \\
    
    \hline
     Group 2   & 67.56 (4.23) & 70.94 (3.86) & 69.54 (4.84) & 66.42 (3.76) & 61.48 (5.15) & 60.33 (4.24) & 60.78 (3.60) & 58.42 (4.08) \\
    
    \hline
    Group 3  & \textbf{71.40 (3.59)} & \textbf{73.52 (3.53)} &  \textbf{72.60 (4.23)}  & \textbf{72.18 (4.01)} & \textbf{62.93 (4.14)} & \textbf{64.63 (3.65)} & \textbf{63.03 (4.36)} & \textbf{63.23 (5.06)}  \\
    \hline

    \end{tabular}%
    
   }
  \label{tab:mean_graph}%
\end{table*}%

\begin{table*}[h!]
  \centering
  \caption{ Initial data labelling: performance comparison of different GNNs sampling strategies based on graph structure. Results are averaged among five trials. For each dataset, we select 0.5\% of the total data as the initial set and evaluate on the rest of the vertices.  Each entry corresponds to ``test accuracy (ACC)| maximum discrepancy (MD)''. $\uparrow$ means the higher the better and $\downarrow$  means the lower the better. {\bf Bold letter} indicates the best sampling strategy for the given dataset and GNN model.
  }  
  \resizebox{\textwidth}{!}
  {
   \setlength\tabcolsep{3pt}
    \begin{tabular}{c|cccc|cccc}
    \hline
              &        \multicolumn{4}{c}{GCN}              & \multicolumn{4}{c}{Sage}            \\
    \hline
          &   Cora & CiteSeer & CoraFull & Coauthor CS & Cora & CiteSeer & CoraFull & Coauthor CS\\
    \hline 
    &  ACC($\uparrow$)|MD($\downarrow$)   &  ACC($\uparrow$)|MD($\downarrow$)  & ACC($\uparrow$)|MD($\downarrow$)  & ACC($\uparrow$)|MD($\downarrow$)    & ACC($\uparrow$)|MD($\downarrow$)  &   ACC($\uparrow$)|MD($\downarrow$)  & ACC($\uparrow$)|MD($\downarrow$)  & ACC($\uparrow$)|MD($\downarrow$)  \\
    \hline
    R   & 37.08|13.23   &  36.35|10.23 & 8.13|3.23 & 57.75|16.23   & 37.69|12.03 &   37.24|11.20 & 8.56|4.20 & 49.72|15.52 \\
    Degree  & 40.87|12.52  &  36.27|9.28 & 10.68|2.32 & 51.50|18.23     & 38.04|11.87 &   35.26|12.05 &   11.56|3.02 & 42.78|13.02   \\
    Centrality & 42.75|10.41  &  38.45|8.15 & 9.49|2.08 & 53.48|16.11     & 42.18|10.15 &   40.81|9.05 &   12.56|3.08 & 45.65|10.11   \\
    PageRank & 47.87|9.52  &  42.27|8.25 & 12.48|2.10 & 55.50|15.23     & 50.15|9.14 &   43.26|9.15 &   13.15|2.98 & 55.16|10.74   \\
    OURS  & \textbf{49.03|8.23 }  &  \textbf{50.98|7.32} & \textbf{15.25|1.23} & \textbf{61.75|14.32}  & \textbf{58.57|8.29}  &  \textbf{48.80|7.03} & \textbf{14.76|2.01} & \textbf{61.63|8.82}    \\
      \hline 
     \hline
         & \multicolumn{4}{c}{GAT} & \multicolumn{4}{c}{GCNII} \\
        \hline
          &   Cora & CiteSeer & CoraFull & Coauthor CS & Cora & CiteSeer & CoraFull & Coauthor CS\\
              \hline 
    &  ACC($\uparrow$)|MD($\downarrow$)   &  ACC($\uparrow$)|MD($\downarrow$)  & ACC($\uparrow$)|MD($\downarrow$)  & ACC($\uparrow$)|MD($\downarrow$)    & ACC($\uparrow$)|MD($\downarrow$)  &   ACC($\uparrow$)|MD($\downarrow$)  & ACC($\uparrow$)|MD($\downarrow$)  & ACC($\uparrow$)|MD($\downarrow$)  \\
       \hline 
       R  &  49.36|15.78 & 43.53|16.27 & 9.55|3.07 & 43.25|15.20 &  31.53|14.02 & 37.29|17.32 & 10.98|3.06 & 52.61|17.02 \\
       Degree  & 54.08|13.52 & 44.65|15.17 & 10.50|2.78 & 47.30|13.04 & 32.19|13.86 & 38.88|16.98 & 10.40|2.79 & 54.70|15.96\\
       Centrality & 56.81|11.52  &  48.15|13.08 & 11.15|2.52 & 48.50|13.18  & 32.89|12.57 &   39.98|16.05 &   12.48|2.52 & 56.88|15.18   \\
    PageRank & 60.91|11.08  &  49.77|12.07 & 12.94|2.11 & 50.50|12.23     & 34.04|12.08 &   42.77|13.95 &   14.56|2.19 & 58.78|14.08   \\
       OURS  &   \textbf{62.76|10.54} & \textbf{51.11|11.07} & \textbf{13.39|1.88}  & \textbf{51.21|11.02}  & \textbf{35.61|11.02} & \textbf{44.50|13.04} & \textbf{15.19|1.96}  & \textbf{60.76|13.18} \\
    \hline
    \end{tabular}%
   }
  \label{tab:init_label}%
\end{table*}%

\subsection{Application in Cold Start Problem for Graph Active Learning}
Active learning is a type of machine learning where the model is actively involved in the labelling process of the training data. In active learning, the model is initially trained on a small labelled dataset and then used to select the most informative examples from the unlabeled dataset to be labelled next. Unsatisfactory initial set selection can lead to slow learning progress and unfair performance, which is referred to as the cold start problem~\cite{cold_start}. Effective resolution of this issue entails an initial set selection that adheres to the following criteria: 1) exclusively leveraging graph structure; 2) achieving robust generalization; and 3) ensuring fairness (with minimal discrepancy in generalization performance) across distinct subgroups. We illustrate how our derived findings can be instrumental in the selection of an effective initial labelled dataset.

Our empirical results show that the selected GNNs are highly distance-aware, meaning that their generalization performance is closely related to the coverage of the training set. To ensure that the model is fair and effective, the initial labelled set should consist of vertices that are in close proximity to the rest of the input graph (in graph distance), as illustrated in Fig.~\ref{fig:different_train_set}. This can be formulated into the following optimization problem:
\begin{equation}\label{eq:max_cover}
\begin{split}
    	& \min_{\vertexSet_0 \subset \mathcal{V}} D_{\mathrm{spd}}(\vertexSet \backslash \vertexSet_0,\vertexSet_0)  \\
		& s.t. ~~~ |\vertexSet_0| = k
\end{split}
\end{equation}
\noindent where $k$ is the given initial set size and $D_s(.)$ is given in Def.~\ref{def:distance}. Intuitively, Eq.\ref{eq:max_cover} aims to select a subset of vertices that exhibit the closest structural similarity (as defined by the graph distance in this case) to the remaining vertices. This problem corresponds to the well-known k-center problem in graph theory~\cite{approx}, which is NP-hard. While finding the optimal solution may be intractable, efficient heuristics exist for tackling Eq.~\ref{eq:max_cover}. Exploring the best heuristic is beyond the scope of this paper. Instead, for our experiment, we adopt the classic farthest-first traversal algorithm to obtain a solution. We refer to this sample selection procedure as the coverage-based sampling algorithm and provide a detailed description of the method used in the experiment within the supplementary material.


We evaluate the initial set selected by our coverage-based sampling algorithm, comparing it to the existing baselines: (1) {\bf R}: uniform random sampling~\cite{al_survey}, and importance sampling with graph properties (2) {\bf Degree}~\cite{hu2020graph}, (3) {\bf Centrality}, (4) {\bf Density} and (5) {\bf PageRank}~\cite{cai2017active}. We apply each strategy to select an initial dataset of 0.5\% of the total data for labelling (training). We evaluate the performance of the learnt model with the rest of the vertices on two metrics: 1) classification accuracy, which reflects the generalization performance, and 2) the maximum discrepancy among structural groups~\cite{fair_metric,fair_gn_survey}, which reflects the fairness. Let $R_i$ denote the mean classification accuracy of the vertex group that is $i$-hop away from $\vertexSet_0$. Then, the maximum discrepancy is computed as $\max \{|R_i-R_j|  | i,j \in 1,...,r \}$, where $r$ is the maximum hop number. We use $r=5$ to keep consistent with other experiments. The results in Table~\ref{tab:init_label} show that coverage-based sampling achieves a significant improvement in model performance and fairness over the other strategies across different datasets, giving a better kick-start to the active learning process of GNNs.

\section{Conclusion and Discussion}\label{sec:discussion_conclusion}
In this work, we study and investigate GNNs which are fundamental to many computer vision and machine learning tasks. In particular, we explore the relationship between the topology awareness and generalization performance of GNNs. We introduce a novel framework that connects the structural awareness of GNNs with approximate metric embedding, offering a fresh perspective on their generalization capabilities in semi-supervised node classification tasks. This structure-agnostic framework facilitates a deeper, more intuitive comprehension of GNN generalizability across varied graph structures. Through a case study centered on graph distance, we demonstrate that our theoretical findings regarding structural results are reflected in the practical generalization performance of GNNs. Additionally, our findings shed light on the cold start problem in graph active learning and could influence various significant GNN applications.

\subsection{Limitation and Future Works}
Our proposed framework introduces a novel angle on GNN generalization performance, yet it is not without limitations. The current approach interprets GNN embeddings in terms of metric mapping and describes the structure awareness of GNNs through distortion. This perspective, however, overlooks the specific dynamics that contribute to reduced distortion within GNNs. Investigating these underlying dynamics would significantly enrich this area of research. Moreover, our study primarily examines the transductive setting. Expanding this analysis to the inductive setting—where the model is trained on one graph but applied to another—would offer a more comprehensive view of GNN generalization capabilities.

{\bf Acknowledgement. }
We would like to thank the anonymous reviewers for their helpful comments. JS and CW are supported by grants from Hong Kong RGC under the contracts HKU 17207621, 17203522 and C7004-22G (CRF).

\newpage
\bibliographystyle{splncs04}
\bibliography{reference}

\newpage
\appendix
\section{Proof of Theorem~\ref{theorem:structural_relation}}\label{appendix:structural_relation}
In this appendix, we provide proof for Theorem~\ref{theorem:structural_relation}.

\begin{proof}
Let $\mathcal{G} = (\mathcal{V}, \mathcal{E}, \featureSet)$ be a given graph and $\vertexSet_0 \subset \vertexSet$ be the training set. Let $s$ be the structure of interest with distance measure $d_s$ and $\vertexSet_i$ be an arbitrary subgroup. Let $\gnnModel$ be a given GNN model with distortion $\alpha$ and scaling factor $r$ and its prediction function $g$.

Let $\sigma:\vertexSet_i \mapsto \vertexSet_0$ denote a mapping that map a vertex $v$ from $\vertexSet_i$ to the closest vertex in $\vertexSet_0$, i.e.,
\begin{equation}
     \sigma(v) = \arg\min_{u \in \vertexSet_0} d_{s}(u,v).
\end{equation}

Let's consider the loss of vertex $v$, $\loss(g \circ \gnnModel(v), y_v)$. By assumption, $\loss$ is smooth and let's denote $B_{\loss}^{up1}$ to be the upper bound for the first derivative with respect to $g \circ \gnnModel(v)$ and $B_{\loss}^{up2}$ to be the upper bound for the first derivative with respect to $y_v$. Then, consider the Taylor expansion of $\loss(g \circ \gnnModel(v), y_v)$ with respect to $\sigma(v)$, which is given as follows
\begin{equation}
\begin{split}
    \loss(g \circ \gnnModel(v), y_v)  \leq \loss(g \circ \gnnModel(\sigma(v)), y_{\sigma(v)}) +  B_{\loss}^{up1}|| g \circ \gnnModel(v) - g \circ \gnnModel(\sigma(v))||  + B_{\loss}^{up2}||y_v - y_{\sigma(v)}||
\end{split}
\end{equation}

Next, let's examine the inequality above term by term and start with $|| g \circ \gnnModel(v) - g \circ \gnnModel(\sigma(v))||$. Let's denote $h_v$ and $h_{\sigma(v)}$ the embedding for vertex $v$ and $\sigma(v)$. Then, we have that,
\begin{equation}
    || g \circ \gnnModel(v) - g \circ \gnnModel(\sigma(v))|| = || g(h_v) - g( h_{\sigma(v)})||
\end{equation}
By definition of distortion as given in Def.~\ref{def:distortion}, we have that,
\begin{equation}
    ||h_v - h_{\sigma(v)}|| \leq r \alpha d_s(v, \sigma(v))
\end{equation}
By assumption, the prediction function $g$ is smooth. Let $B_{g}$ denote the upper bound of first derivative of the prediction function $g$. Then, we have 
\begin{equation}
    || g(h_v) - g(h_{\sigma(v)})|| \leq B_g^{up} ||h_v - h_{\sigma(v)}||
\end{equation}
Substitute all these back, we have
\begin{equation}
     || g \circ \gnnModel(v) - g \circ \gnnModel(\sigma(v))|| \leq B_g^{up} r \alpha d_s(v, \sigma(v))
\end{equation}
Next, let's consider $||y_v - y_{\sigma(v)}||$. Similarly, by the data smoothness assumption and distortion, we have that 
\begin{equation}
    ||y_v - y_{\sigma(v)}|| \leq B_l ||h_v - h_{\sigma(v)}|| \leq B_l^{up} r \alpha d_s(v, \sigma(v))
\end{equation}

Substitute these back to the inequality we start with, we have that
\begin{equation}
    \begin{split}
            \loss(g \circ \gnnModel(v), y_v) & \leq \loss(g \circ \gnnModel(\sigma(v)), y_{\sigma(v)}) + 
             B_{\loss}^{up1}B_g^{up} r \alpha d_s(v, \sigma(v)) + B_{\loss}^{up2}B_l^{up} r \alpha d_s(v, \sigma(v)) \\
           & \leq \loss(g \circ \gnnModel(\sigma(v)), y_{\sigma(v)}) +                 
           (B_l^{up}B_{\loss}^{up1}+B_g^{up}B_{\loss}^{up2}) r \alpha \max_{v' \in \vertexSet_i} d_s(v', \sigma(v'))\\
           & = \loss(g \circ \gnnModel(\sigma(v)), y_{\sigma(v)}) + 
           (B_l^{up}B_{\loss}^{up1}+B_g^{l}B_{\loss}^{up2}) r \alpha D_{s}(\vertexSet_i, \vertexSet_0).
    \end{split}
\end{equation}

Since the inequality above holds for every vertex $v$, then we have 

\begin{equation}
    \begin{split}
         R^{\loss}(g \circ \gnnModel,\vertexSet_i) & \leq R^{\loss}(g \circ \gnnModel,\vertexSet_0) + 
         (B_l^{up}B_{\loss}^{up1}+B_g^{up}B_{\loss}^{up2})  r \alpha D_{s}(\vertexSet_i, \vertexSet_0)\\
         & = R^{\loss}(g \circ \gnnModel,\vertexSet_0) + \mathcal{O} (\alpha  D_s(\vertexSet_i,\vertexSet_0))
    \end{split}
\end{equation}

\end{proof}
\section{Proof of Theorem~\ref{theorem:sub_group_performance}}\label{appendix:trainset_proof}
Before diving into the detailed proof, we present an outline of the structure of the proof and prove a lemma which we use in the proof of the theorem.

{\bf Outline of the proof for the thereom}
	\begin{enumerate}
		\item Suppose we are given $\vertexSet_i$ and $\vertexSet_j$, two test groups which satisfy the premise of the theorem;
		\item Then, we can approximate and bound the loss of each vertex in these groups based on the nearest vertex in the training set by extending the result from Theorem~\ref{theorem:structural_relation};
		\item If we can show that there exists a constant independent of the property of each test group, then we obtain the results of Theorem~\ref{theorem:sub_group_performance}.
	\end{enumerate}

\begin{proof}
Let $\mathcal{G} = (\mathcal{V}, \mathcal{E}, \featureSet)$ be a given graph and $\vertexSet_0 \subset \vertexSet$ be the training set. Let $s$ be the structure of interest with distance measure $d_s$ and $\vertexSet_i$ be an arbitrary subgroup. Let $\gnnModel$ be a given GNN model with distortion $\alpha$ and scaling factor $r$ and its prediction function $g$. 

	
Let $\vertexSet_i$ and $\vertexSet_j$ be the two test groups satisfy the premise of the theorem.





Let $Q_i: \vertexSet_i \mapsto \vertexSet_0$ be the mapping that maps vertex $u \in \vertexSet_i$ to the closest vertex $v \in \vertexSet_0$.  For simplicity, we denote the vertex as $q_u = Q_i(u)$. We can do a linear approximation of $u$ around $q_u$ and obtain a similar upper bound in Theorem~\ref{theorem:structural_relation}, which is given as follows:

\begin{equation}\label{eq:upper}
\begin{split}
    R^{\loss}(g \circ \gnnModel,\vertexSet_i)  \leq  R^{\loss}(g \circ \gnnModel,\vertexSet_0) + (B_l^{up}B_{\loss}^{up1}+B_g^{up}B_{\loss}^{up2})  r \alpha D_{s}(\vertexSet_i, \vertexSet_0)
\end{split}
\end{equation}






Similarly, let $Q_j: \vertexSet_j \mapsto \vertexSet_0$ be the mapping that maps vertex $u \in \vertexSet_j$ to the closest vertex $u \in \vertexSet_0$. For simplicity, we denote the vertex as $q'_u = Q_j(u)$. Because we assume the training loss is arbitrarily small (easily obtainable with sufficient parameters) and the groups are in the local neighbourhood of the training set, we can do a linear approximation of $u$ around $q'_u$ and have the following lower bound:

\begin{equation}
\begin{split}
    \loss(g \circ \gnnModel(v), y_v) & \geq \loss(g \circ \gnnModel(q_v), y_{q_v}) +  B_{\loss}^{l1}|| g \circ \gnnModel(v) - g \circ \gnnModel(q_v)||  + B_{\loss}^{l2}{\loss}||y_v - y_{q_v}|| \\
    & \geq \loss(g \circ \gnnModel(q_v), y_{q_v}) + 
  (B_l^{l}B_{\loss}^{l1}+B_g^{l}B_{\loss}^{l2}) r  D_{s}(\vertexSet_i, \vertexSet_0).
\end{split}
\end{equation}
where $B_{\loss}^{l1}$ and $B_{\loss}^{l1}$ are the lower bound of the $\loss(a,b)$ function with respect to $a$ and $b$. $B_l^{l}$ and $B_g^{l}$ are the lower bound for the prediction function and data generation. $B_{l1}$, $B_{l2}$,$B_l^{l}$,$B_g^{l}$ exists because $\loss, g$ and the data generation are smooth. Then, we have
\begin{equation}\label{eq:lower}
\begin{split}
    R^{\loss}(g \circ \gnnModel,\vertexSet_j) & \geq  R^{\loss}(g \circ \gnnModel,\vertexSet_0) + 
    (B_l^{l}B_{\loss}^{l1}+B_g^{l}B_{\loss}^{l2}) r  D_{s}(\vertexSet_i, \vertexSet_0).
\end{split}
\end{equation}



Now suppose $D_{s}(\vertexSet_i, \vertexSet_0) > D_{s}(\vertexSet_j, \vertexSet_0) $. Let's consider the difference $\vertexSet_i$ and $\vertexSet_j$ with Eq.~\eqref{eq:lower} being applied on $\vertexSet_i$ and Eq.~\eqref{eq:upper} being applied on $\vertexSet_j$, then we have that 
\begin{equation}
    \begin{split}
        R^{\loss}(g \circ \gnnModel,\vertexSet_i) -R^{\loss}(g \circ \gnnModel,\vertexSet_j) & \geq  R^{\loss}(g \circ \gnnModel,\vertexSet_0) +  (B_l^{l}B_{\loss}^{l1}+B_g^{l}B_{\loss}^{l2}) r  D_{s}(\vertexSet_i, \vertexSet_0) -\\
   & [ R^{\loss}(g \circ \gnnModel,\vertexSet_0) + 
    (B_l^{up}B_{\loss}^{up1}+B_g^{up}B_{\loss}^{up2}  r \alpha D_{s}(\vertexSet_j, \vertexSet_0)]
    \end{split}
\end{equation}

Then, we can derive that if we have 
\begin{equation}
    \begin{split}
        (B_l^{l}B_{\loss}^{l1}+B_g^{l}B_{\loss}^{l2}) r  D_{s}(\vertexSet_i, \vertexSet_0) 
        > B_l^{up}B_{\loss}^{up1}+B_g^{up}B_{\loss}^{up2} D_{s} r \alpha (\vertexSet_j, \vertexSet_0)
    \end{split}
\end{equation}
or equivalently,
\begin{equation}
    \begin{split}
          D_{s}(\vertexSet_i, \vertexSet_0) 
        > \frac{B_l^{up}B_{\loss}^{up1}+B_g^{up}B_{\loss}^{up2}}{(B_l^{l}B_{\loss}^{l1}+B_g^{l}B_{\loss}^{l2})}  \alpha D_{s}(\vertexSet_j, \vertexSet_0)
    \end{split}
\end{equation}

 Let $\delta = \frac{B_l^{up}B_{\loss}^{up1}+B_g^{up}B_{\loss}^{up2}}{(B_l^{l}B_{\loss}^{l1}+B_g^{l}B_{\loss}^{l2})}$, we get the result desired by the theorem.

\end{proof}
\section{Procedure for Solving Eq.~\eqref{eq:max_cover}}
First recall that the optimization problem we want to solve is as follows.
\begin{equation}
		 \min_{\vertexSet_0 \subset \mathcal{V}} D_s(\vertexSet \backslash \vertexSet_0,\vertexSet_0)   ~~~
		s.t. ~~~ |\vertexSet_0| = k
\end{equation}

\noindent where $k$ is the given initial set size and $D_s(.)$ is given in Def.~\ref{def:distance} with shortest path distance as the metric. Eq.~\ref{eq:max_cover} amounts to the well-known k-center problem~\cite{approx} in graph theory and is NP-hard. For our experiment, we use Eq.~\ref{eq:max_cover} as guidance and modify the simple greedy algorithm, farthest-first traversal, to be a sampling method for obtaining a solution. We now describe how we modified the standard greedy algorithm into a sampling algorithm. We start with describing the standard greedy algorithm.

\begin{algorithm}[!h]
  \caption{k-center Greedy}
  \label{alg:k-center}
\begin{algorithmic}
  \REQUIRE $\mathcal{V}$ //vertex set of a given graph\\
  \REQUIRE $k$ //number of centers\\
  
  Initialize an empty array $V$
  \begin{enumerate}
     \item randomly picks a random vertex $v$ to $V$
      \item pick the next vertex to be the vertex furthest away from $V$
      \item repeat until $|V| = k$
  \end{enumerate}
\end{algorithmic}
\end{algorithm}

The procedure described above is a standard 2-approximation greedy algorithm for k-center problem. Next, we describe its sampling variant which use the distance as the chance of being sampled.

\begin{algorithm}[!h]
  \caption{Coverage-based Sampling}
  \label{alg:coverage_based_sampling}
\begin{algorithmic}
  \REQUIRE $\mathcal{V}$ //vertex set of a given graph\\
  \REQUIRE $k$ //size of set to be sampled\\

  Sample/select the set of size $k$ with the following procedures:
  \begin{enumerate}
     \item initialize $V$ with the vertex of the highest degree.
      \item sample $u$ from $\mathcal{V}\backslash V$ based on probability $p_u$ proportion to $D_s(u,\vertexSet\backslash V)$, following the formula,
      $$p_u = \frac{D_{s}(u,V)}{\sum_{v \in \vertexSet\backslash V}D_{s}(V,V)}$$
      \item repeat until $k$ vertices are sampled
  \end{enumerate}
\end{algorithmic}
\end{algorithm}

\subsection{Running Time Complexity}
It is obvious that the procedure above runs in $\mathcal{O}(k)$ if the distance between any pair of vertices is given and runs in $\mathcal{O}(kn)$ if the distance needed to be computed for each step, where $n$ is the number of vertice.
\section{Additional Experiments Details and Results}\label{appendix:additional_experiments}
In this appendix, we provide additional experimental results and include a detailed set-up of the experiments for reproducibility.

\subsection{Hardware and Software}
All the experiments of this paper are conducted on the following machine

CPU: two Intel Xeon Gold 6230 2.1G, 20C/40T, 10.4GT/s, 27.5M Cache, Turbo, HT (125W) DDR4-2933

GPU: four NVIDIA Tesla V100 SXM2 32G GPU Accelerator for NV Link

Memory: 256GB (8 x 32GB) RDIMM, 3200MT/s, Dual Rank

OS: Ubuntu 18.04LTS

\subsection{Training Hyper-parameters and Detailed Dataset Description}
Table~\ref{tab:data_description} gives a more detailed description of the statistic of datasets used in the experiment as well as the hyper-parameters used for training. We keep the hyper-parameters the same across different models. To keep the experiment consistent, we do not employ randomized regularization techniques such as dropout. To ensure a fair comparison, we make sure each experiment is trained till convergence. 

\begin{table*}[!h]
  \centering
  \caption{Description on the datasets and hyper-parameter different tasks.
  }  
  {\small
   \setlength\tabcolsep{4pt}
    \begin{tabular}{|c|cc|cc|c|}
    \hline
        Datasets  &        Cora  & Citeseer & Cora Full & Coauthor CS & OGB-Arxiv          \\
            
    \hline
          Number of Vertexes & 2,708 & 3,327 & 19,793 & 18,333 & 169,343\\
          Number of Edges   & 10,556 & 9,228 & 126,842 & 163,788 & 1,166,243	\\
          Dimension of Node Feature & 1,433 & 3,703 & 8,710 & 6,805 & 128 \\
         Number of Class & 7 & 6 & 70 & 15 & 40\\

    \hline
        Number of Hidden Layer & 2 & 2 & 2 & 2  & 2\\
        Hidden Layer Dimension & 128 & 128 & 256 & 256 & 256\\
    \hline
     Prediction Head  & \multicolumn{5}{c}{MLP} \\
     Prediction Head Activation & \multicolumn{5}{c}{ReLU} \\
     Prediction Head Dimension & 32 & 32 & 128 & 128 & 256\\
    \hline
        Optimizer & \multicolumn{5}{c}{Adam} \\
        Learning Rate & \multicolumn{5}{c}{0.01}   \\
        Weight Decay & \multicolumn{5}{c}{0.0005}   \\
        Dropout  & \multicolumn{5}{c}{0} \\
        Epoch   & \multicolumn{5}{c}{100}  \\
    \hline
    \end{tabular}%
   }
  \label{tab:data_description}%
\end{table*}%

\subsection{Additional Experiment Results}
In addition to the datasets we present in the main text, we also conduct experiments on Citeseer, Cora Full and OGBN-arxiv. The results of these experiments show that the theoretical structures proved in this paper are indeed consistent across different datasets. Next, we present additional experimental results on these datasets for validating the main theorems of our papers.

\subsubsection{Graph Distance and Performance}
\begin{figure*}[!h]
\centering
\subfigure[GCN, Citeseer]{
\includegraphics[width=.225\textwidth]{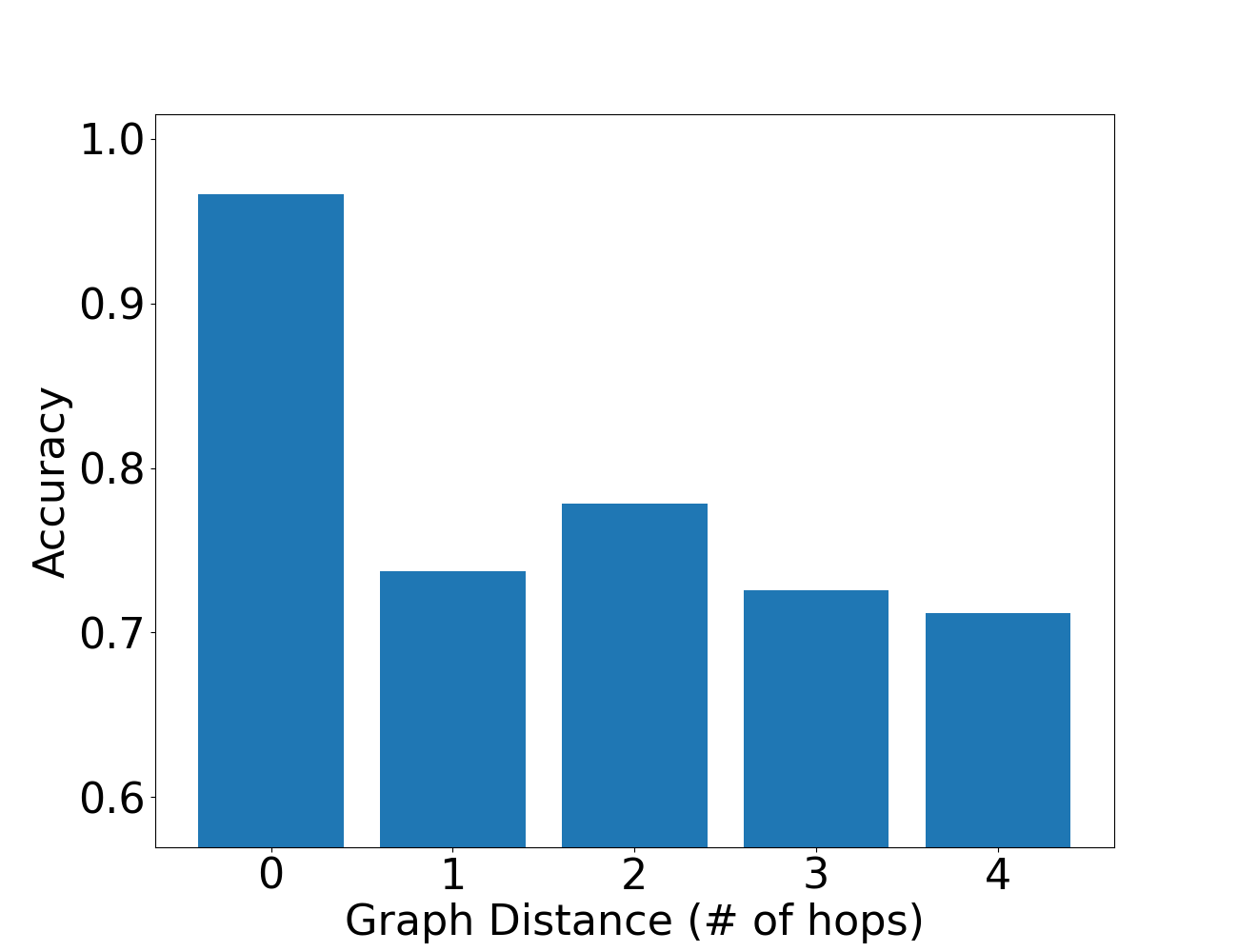}
}
\subfigure[GraphSAGE, Citeseer]{
\includegraphics[width=.225\textwidth]{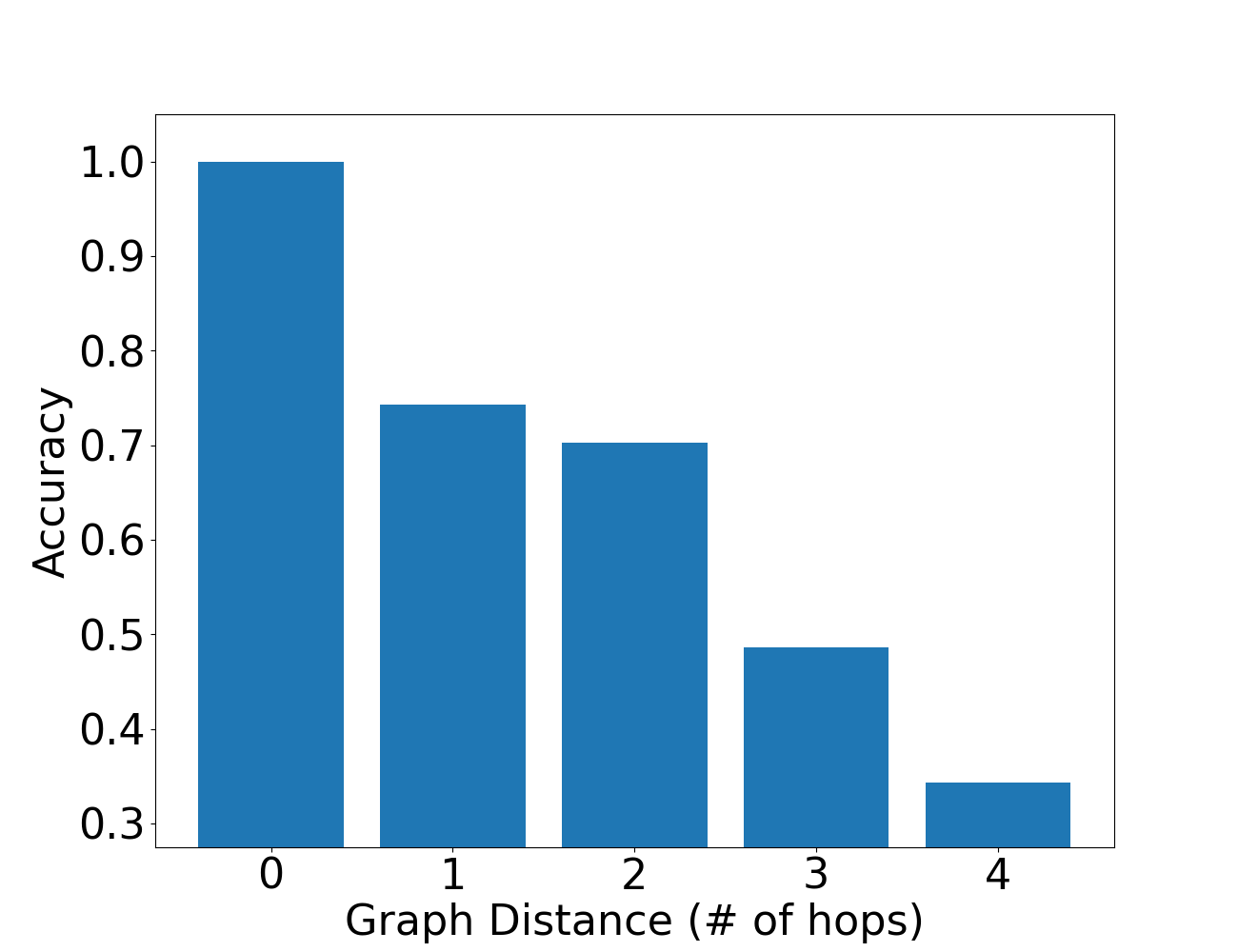}
}
\subfigure[GAT, Citeseer]{
\includegraphics[width=.225\textwidth]{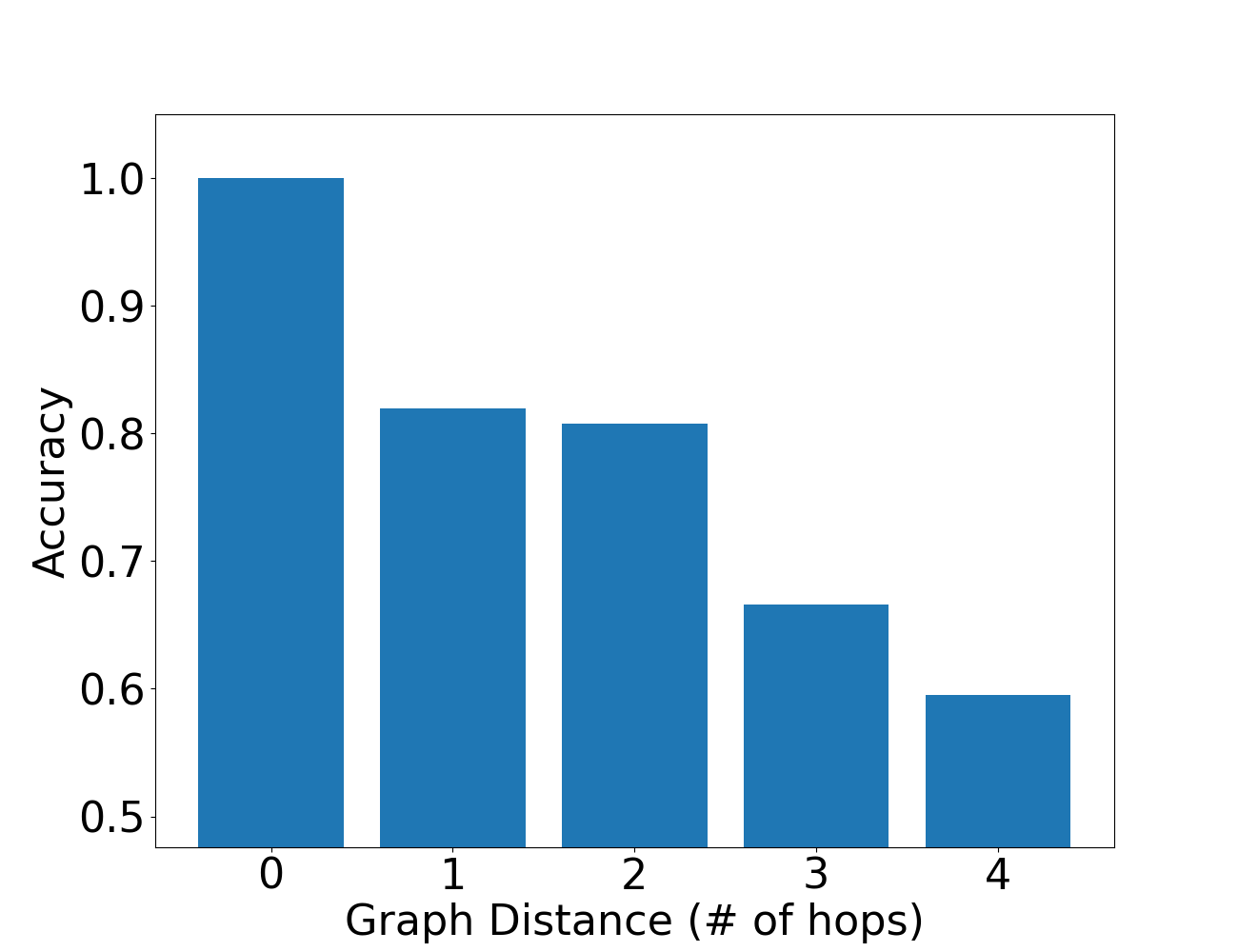}
}
\subfigure[GCNII, Citeseer]{
\includegraphics[width=.225\textwidth]{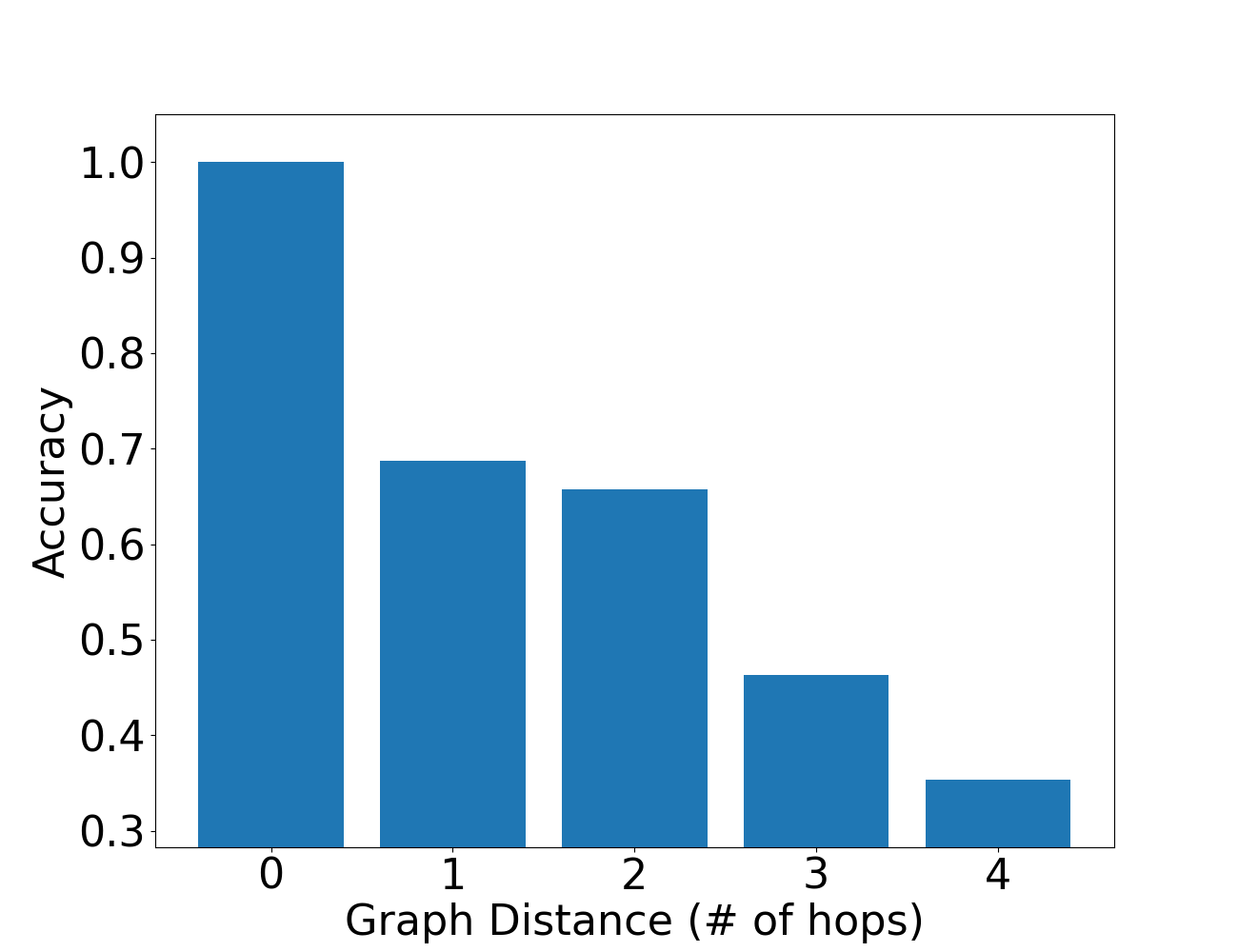}
}
\caption{Graph distance vs.~accuracy. Additional Results on Citeseer}
\label{fig:group_accuracy_citeseer}
\end{figure*}

\begin{figure*}[!h]
\centering

\subfigure[GCN, CoralFull]{
\includegraphics[width=.225\textwidth]{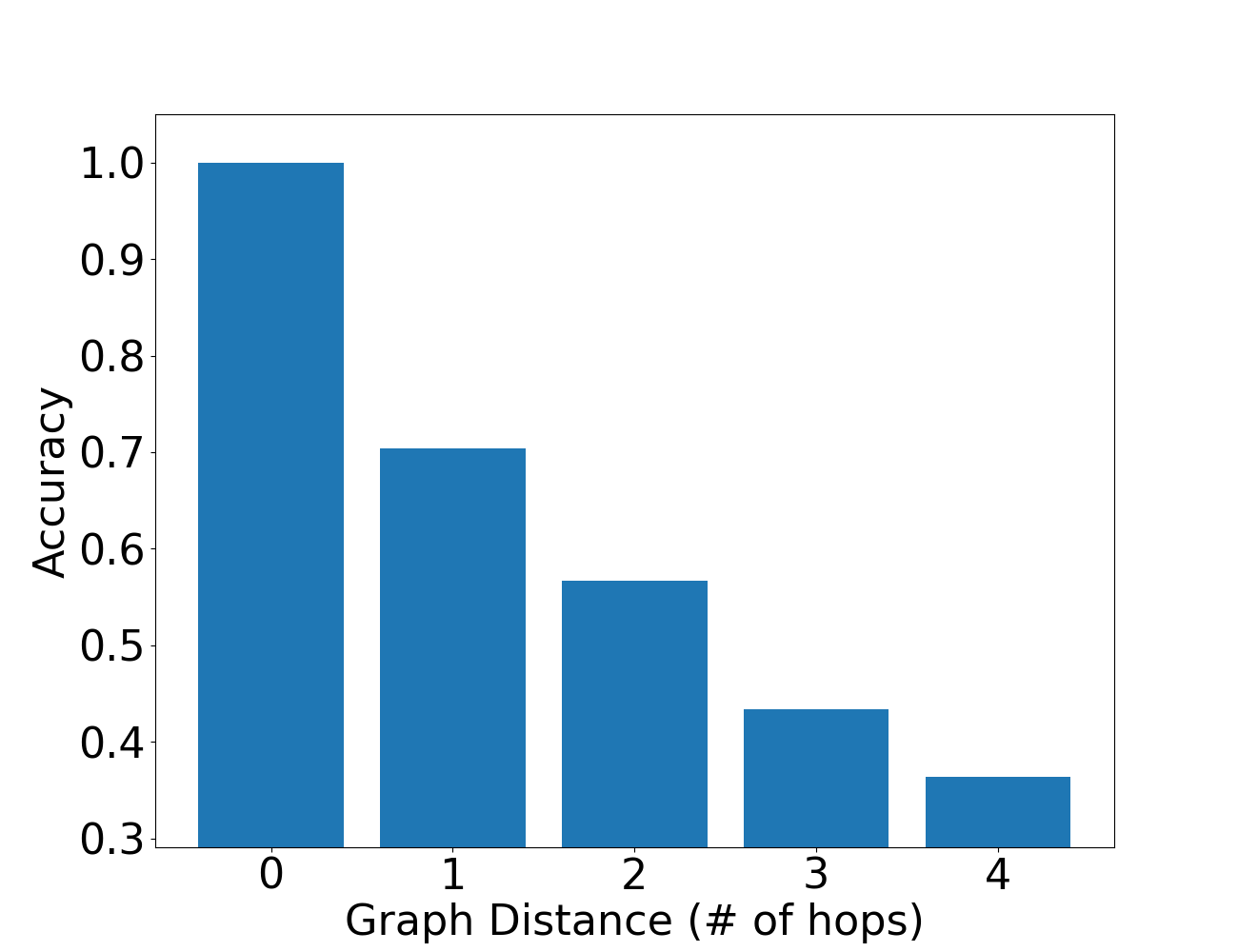}
}
\subfigure[GraphSAGE, CoralFull]{
\includegraphics[width=.225\textwidth]{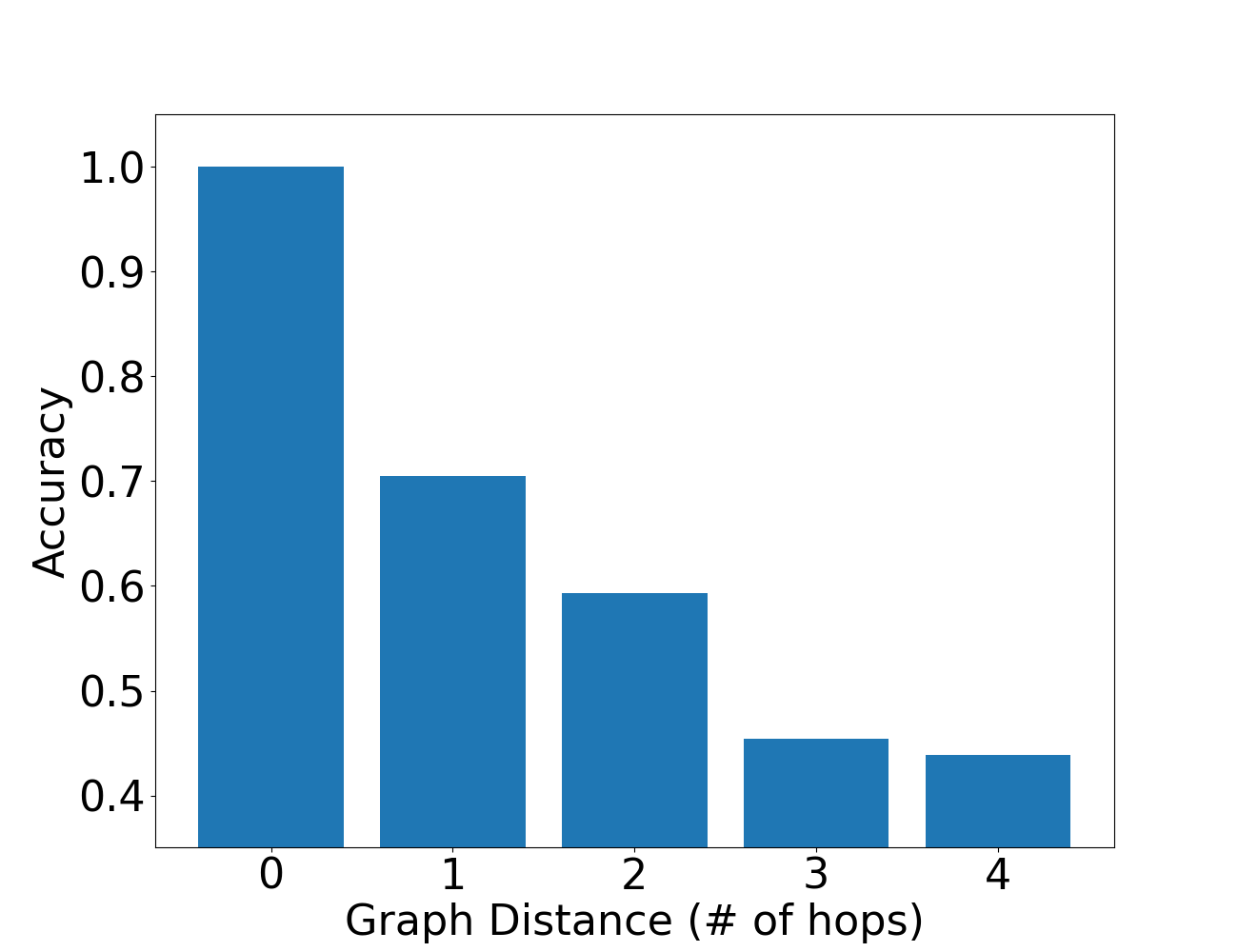}
}
\subfigure[GAT, CoralFull]{
\includegraphics[width=.225\textwidth]{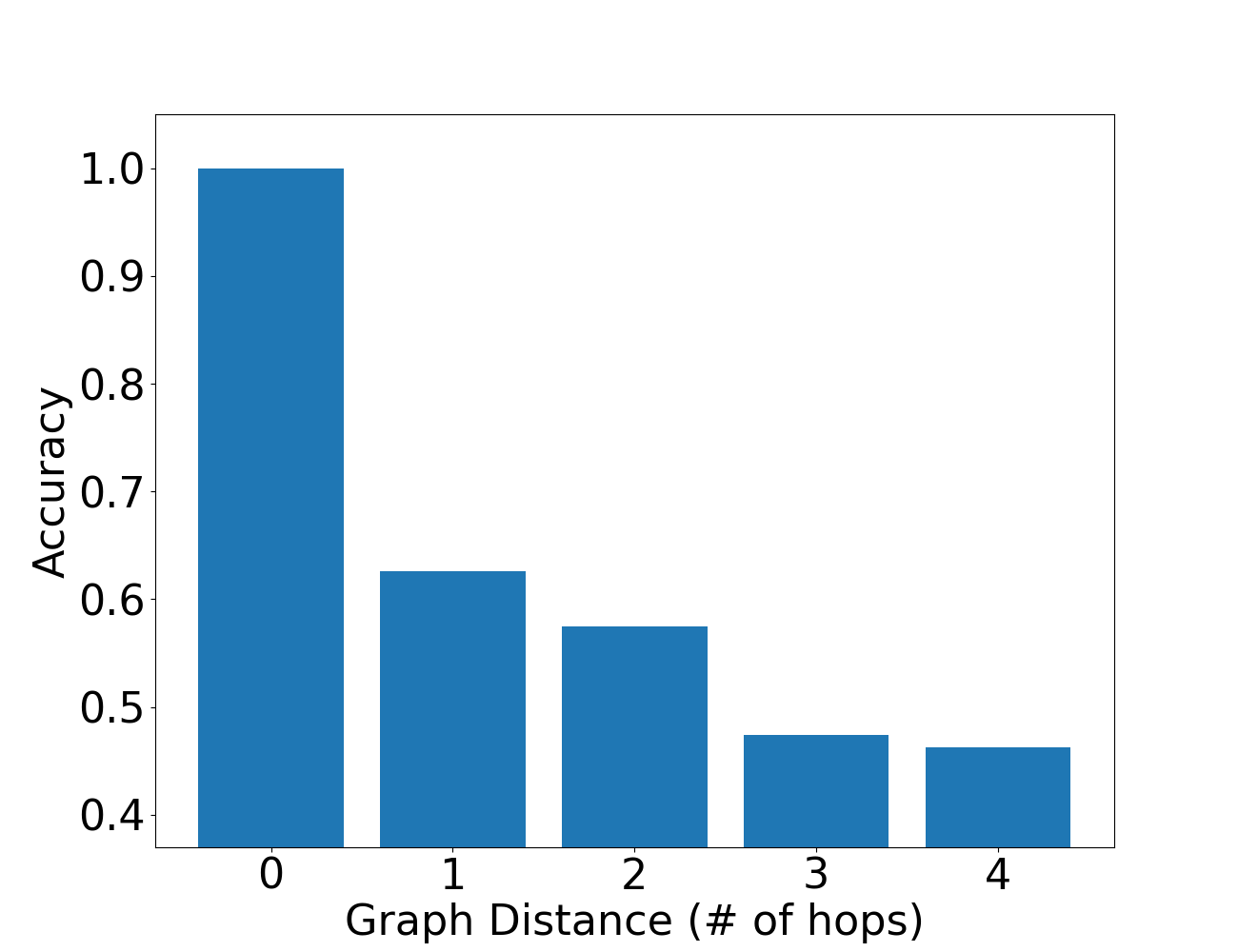}
}
\subfigure[GCNII, CoralFull]{
\includegraphics[width=.225\textwidth]{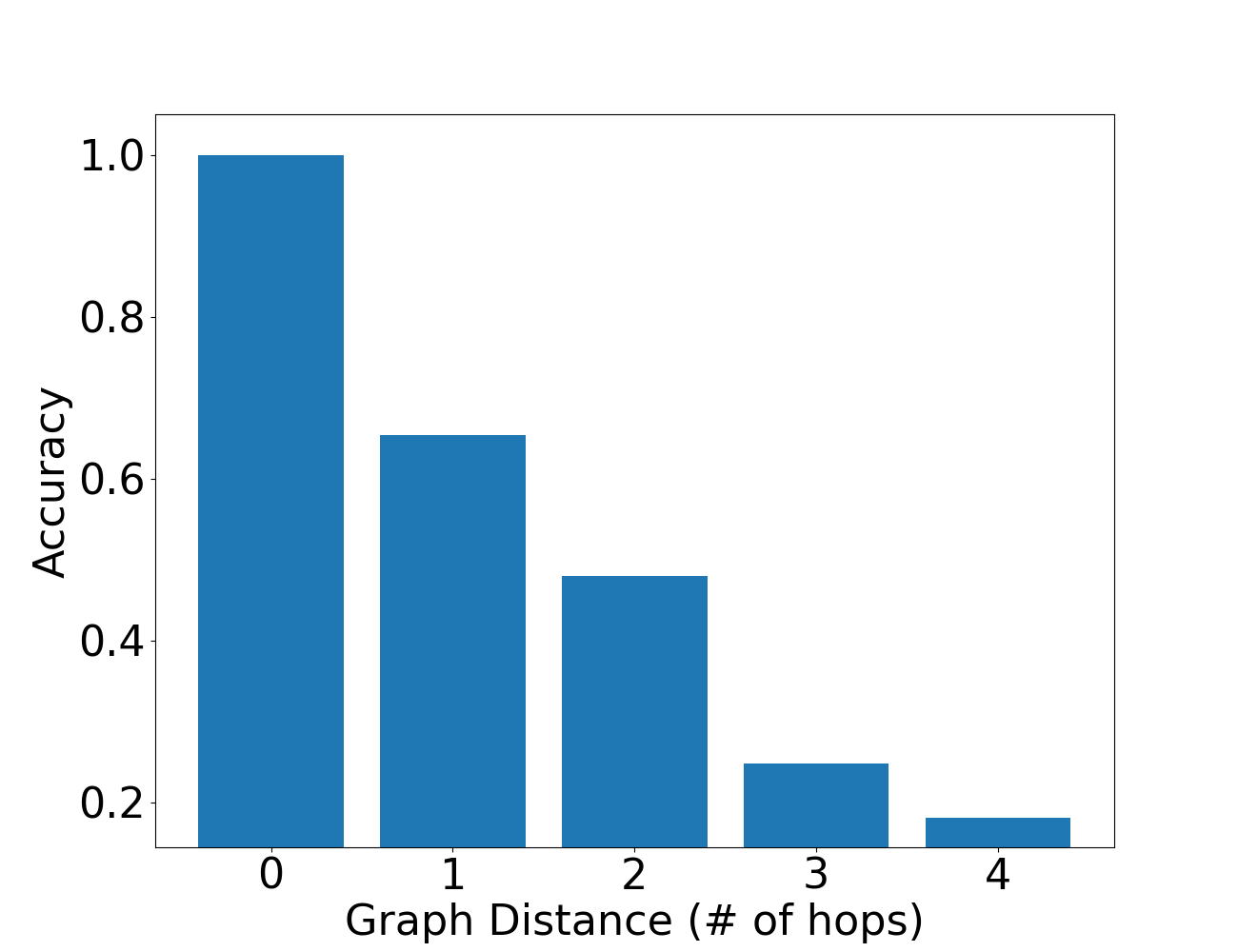}
}
\caption{Graph distance vs.~accuracy. Additional Results on CoraFull}
\label{fig:group_accuracy_citefull}
\end{figure*}

\begin{figure*}[!h]
\centering
\subfigure[GCN, OGB-Arxiv]{
\includegraphics[width=.225\textwidth]{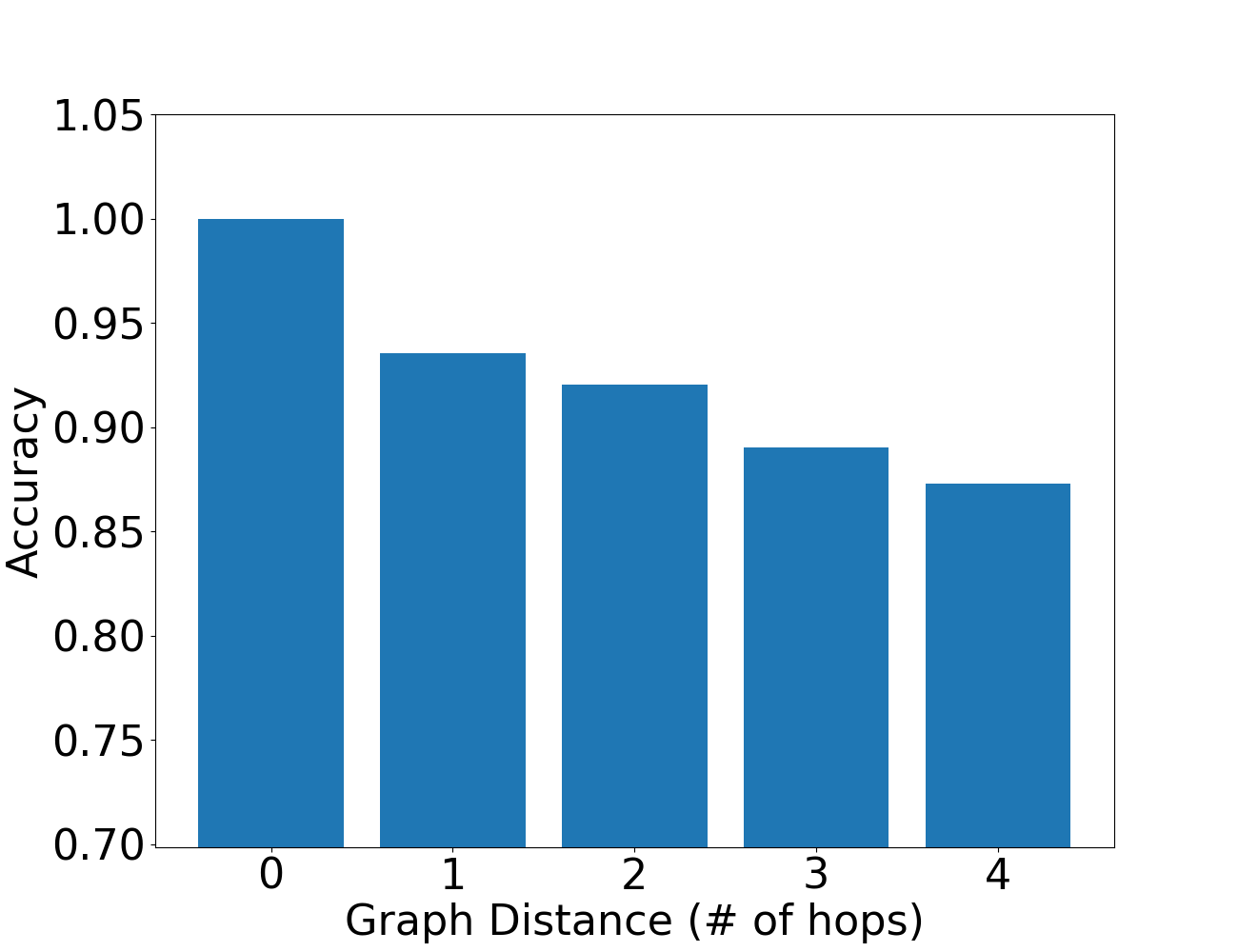}
}
\subfigure[GraphSAGE, OGB-Arxiv]{
\includegraphics[width=.225\textwidth]{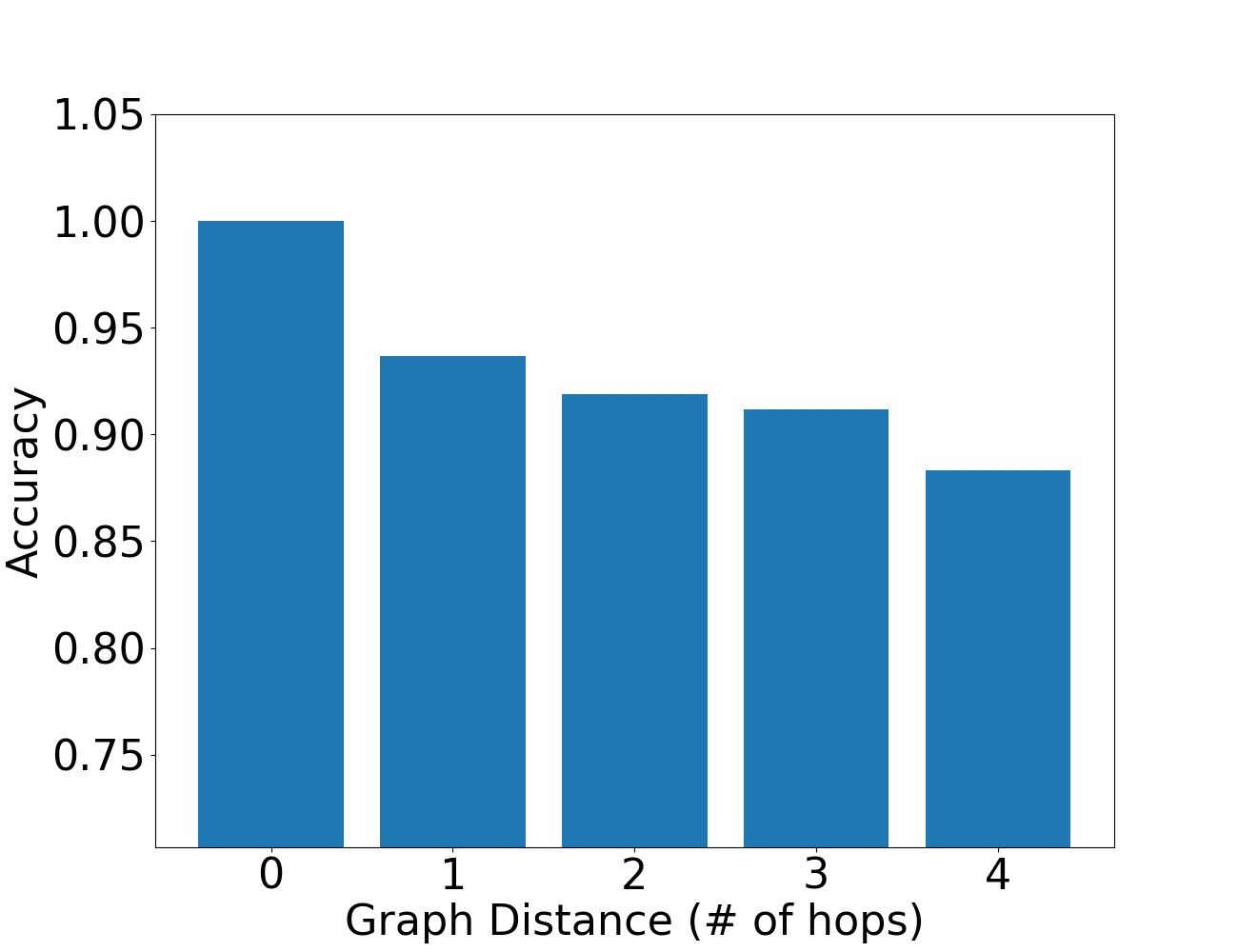}
}
\subfigure[GAT, OGB-Arxiv]{
\includegraphics[width=.225\textwidth]{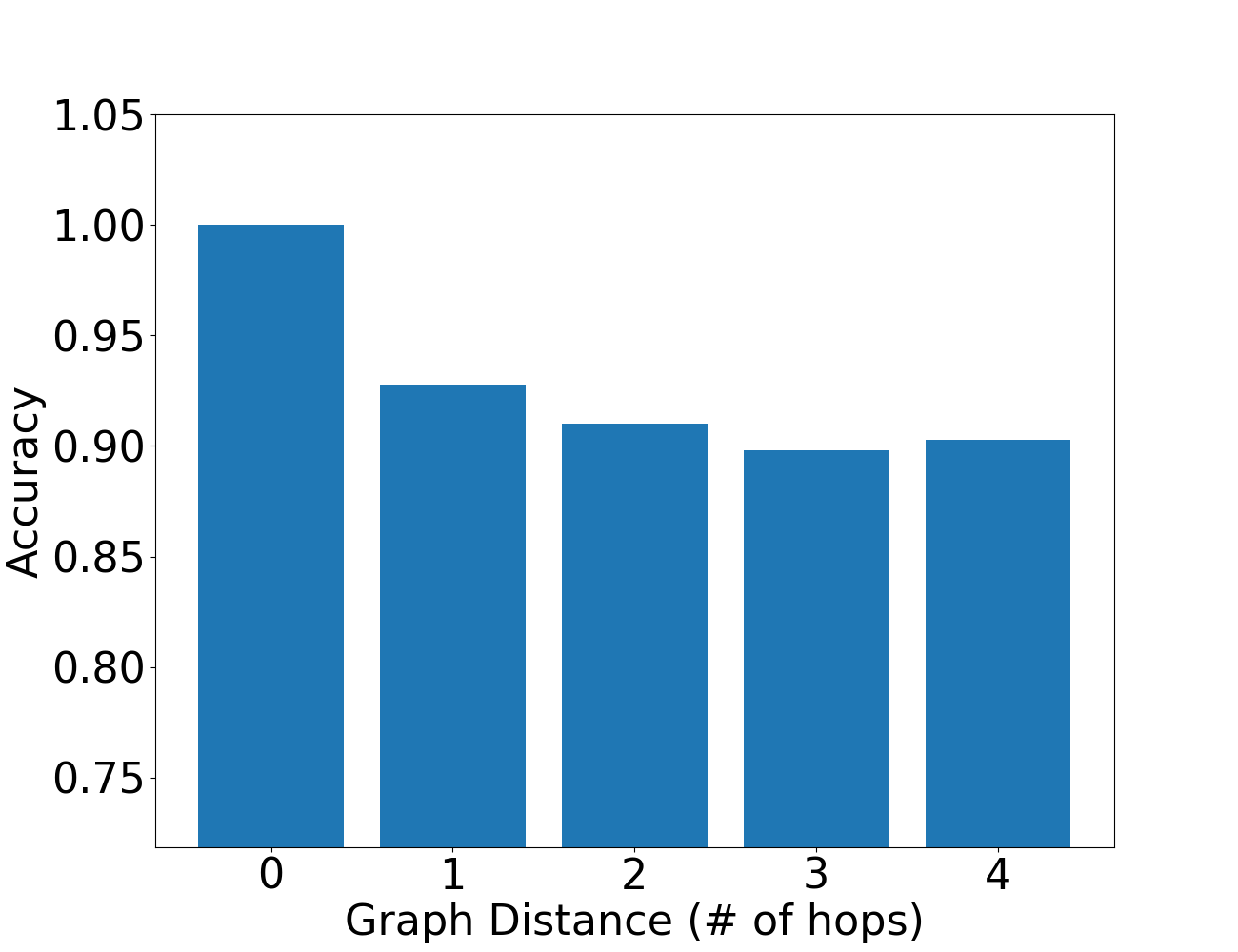}
}
\subfigure[GCNII, OGB-Arxiv]{
\includegraphics[width=.225\textwidth]{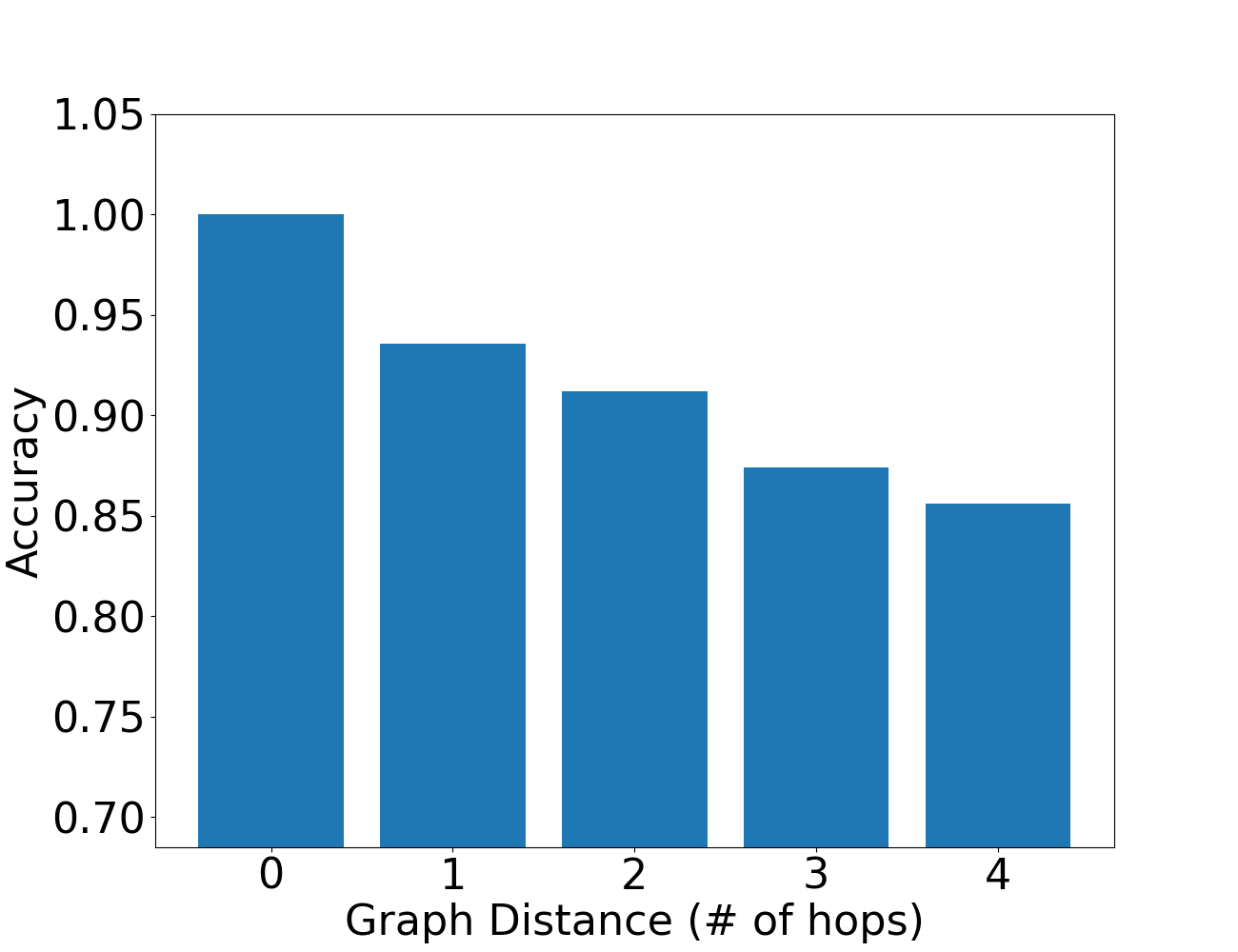}
}
\caption{Graph distance vs.~accuracy. Additional Results on OGBN-Arxiv}
\label{fig:group_accuracy_ogb}
\end{figure*}

 We present additional results with Citeseer, CoraFull, and OGBN-Arxiv on the relation between the graph distance (shortest-path distance). For these experiments, we follow the same setting as the one presented in Fig.~\ref{fig:group_accuracy}. We compute the average prediction accuracy in each group of vertices (corresponding to vertices at each hop of the neighbourhood of the training set). '0 hop' indicates the training set. We observe that GNNs tend to predict better on the
vertices that are closer to the training set. The results are presented in Fig.~\ref{fig:group_accuracy_citeseer}, ~\ref{fig:group_accuracy_citefull} and ~\ref{fig:group_accuracy_ogb}.

\subsubsection{Subgroup Embedding Distance and Accuracy}
\begin{figure*}[!h]
\centering
\subfigure[GCN, Citeseer]{
\includegraphics[width=.225\textwidth]{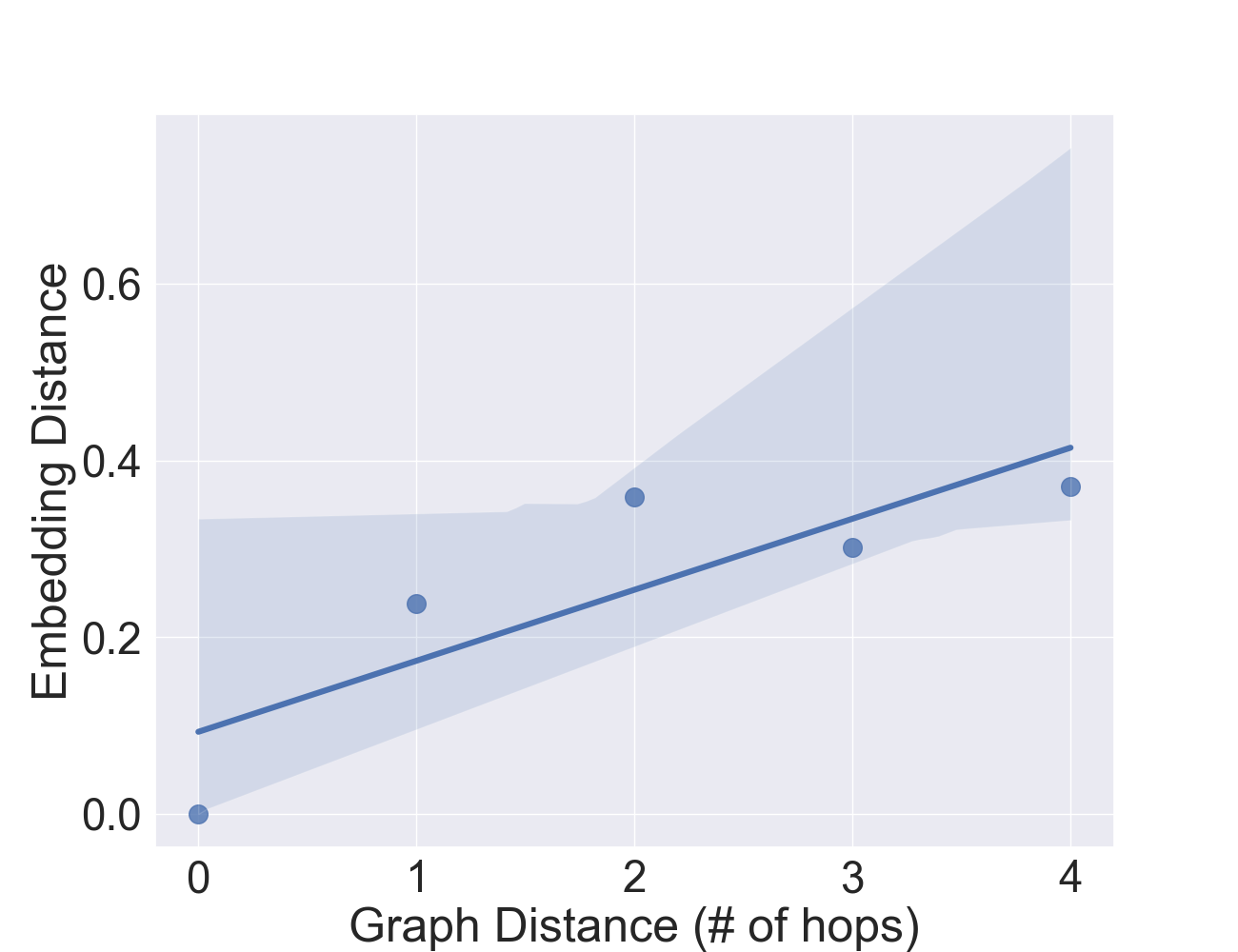}
}
\subfigure[GraphSAGE, Citeseer]{
\includegraphics[width=.225\textwidth]{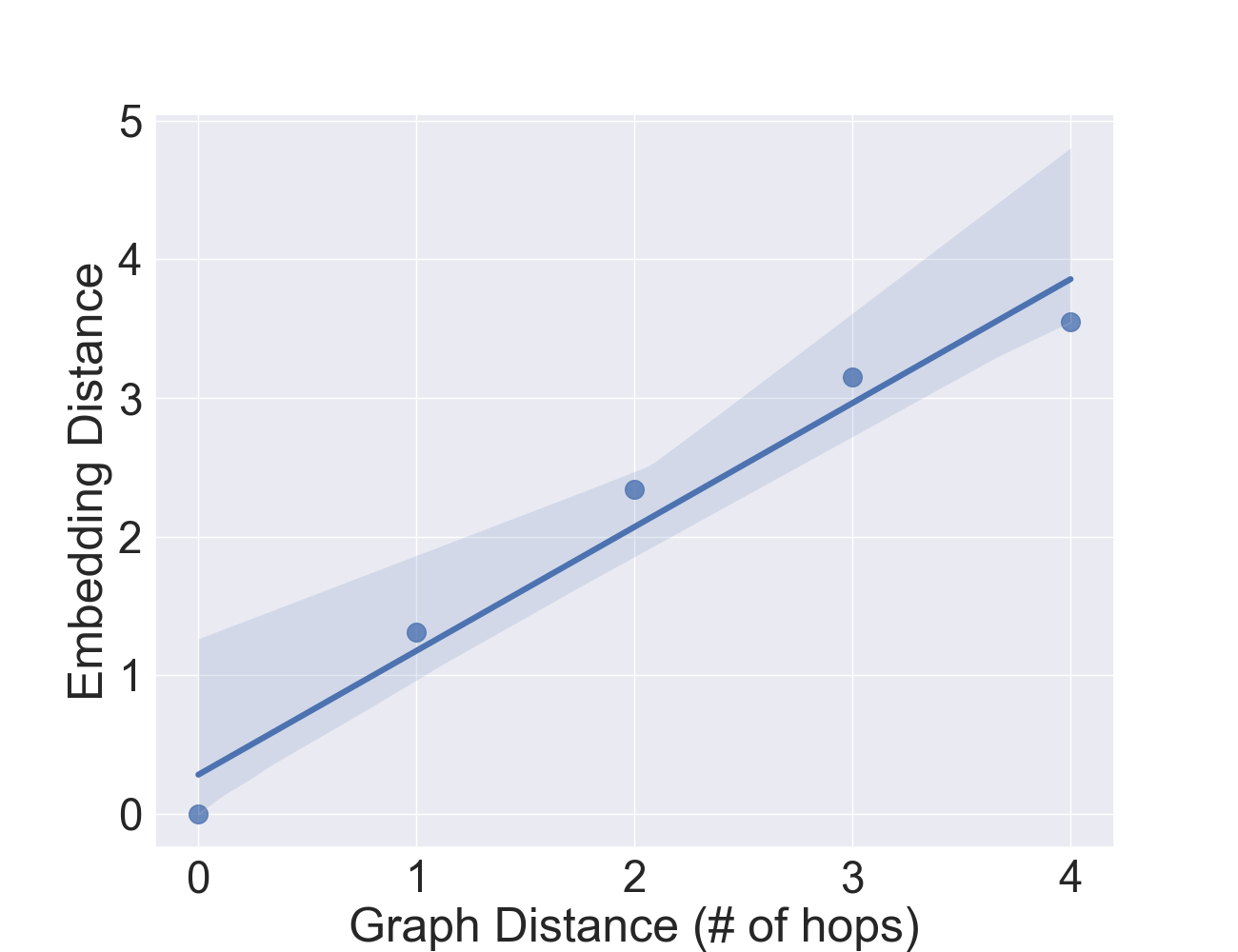}
}
\subfigure[GAT, Citeseer]{
\includegraphics[width=.225\textwidth]{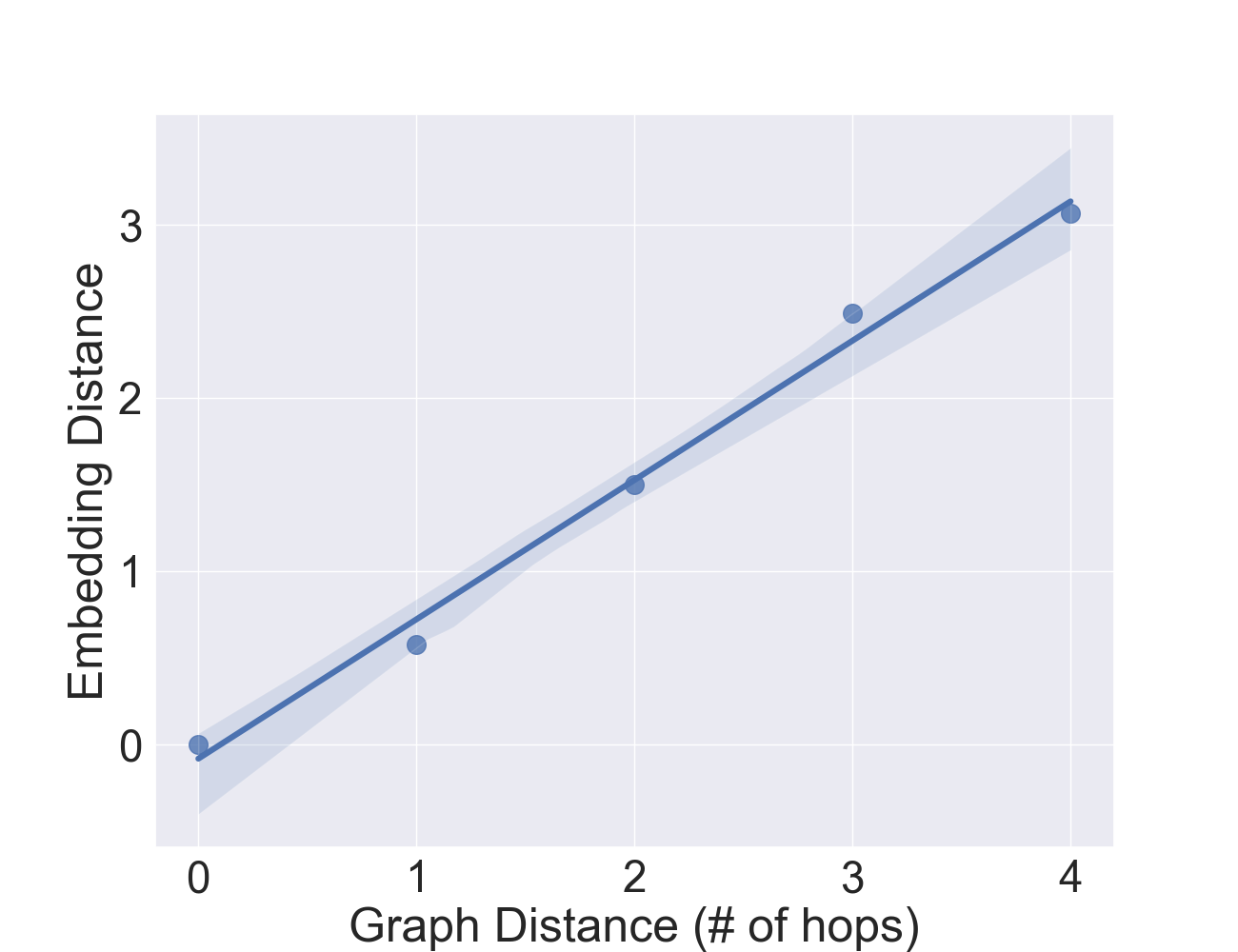}
}
\subfigure[GCNII, Citeseer]{
\includegraphics[width=.225\textwidth]{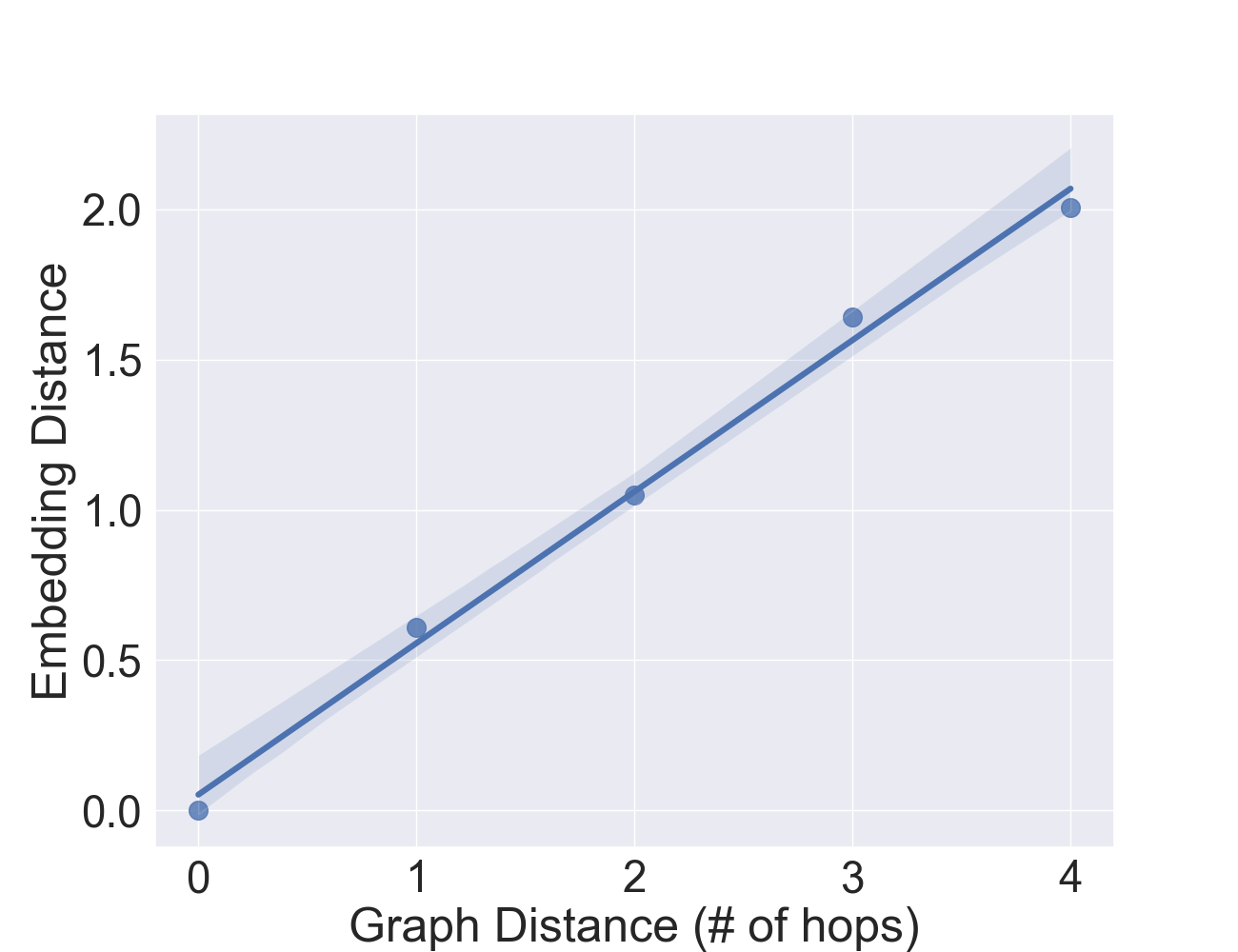}
}
\caption{Graph distance vs.~embedding distance. Additional Results on Citeseer}
\label{fig:group_embedding_citeseer}
\end{figure*}

\begin{figure*}
\centering
\subfigure[GCN, CoraFull]{
\includegraphics[width=.225\textwidth]{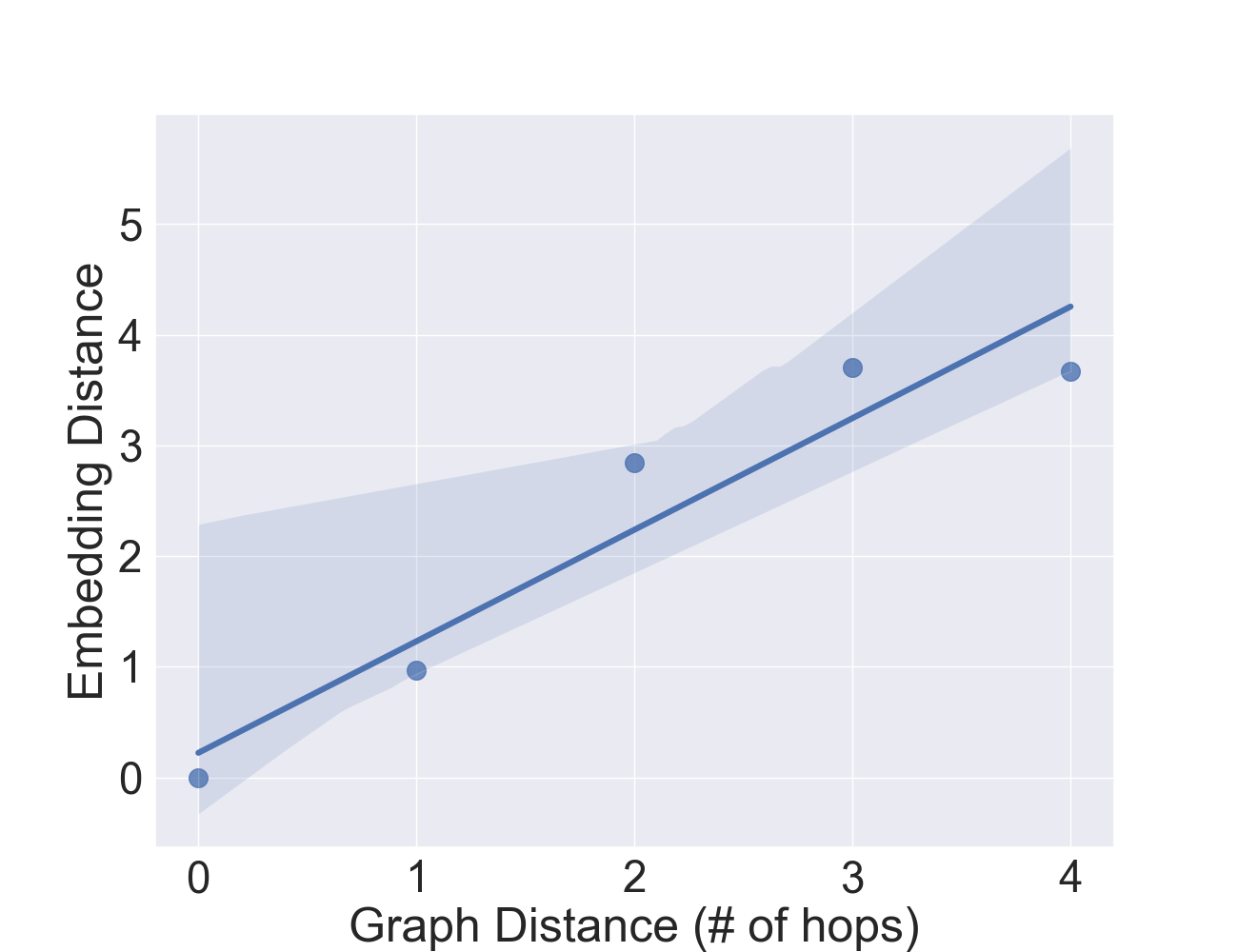}
}
\subfigure[GraphSAGE, CoraFull]{
\includegraphics[width=.225\textwidth]{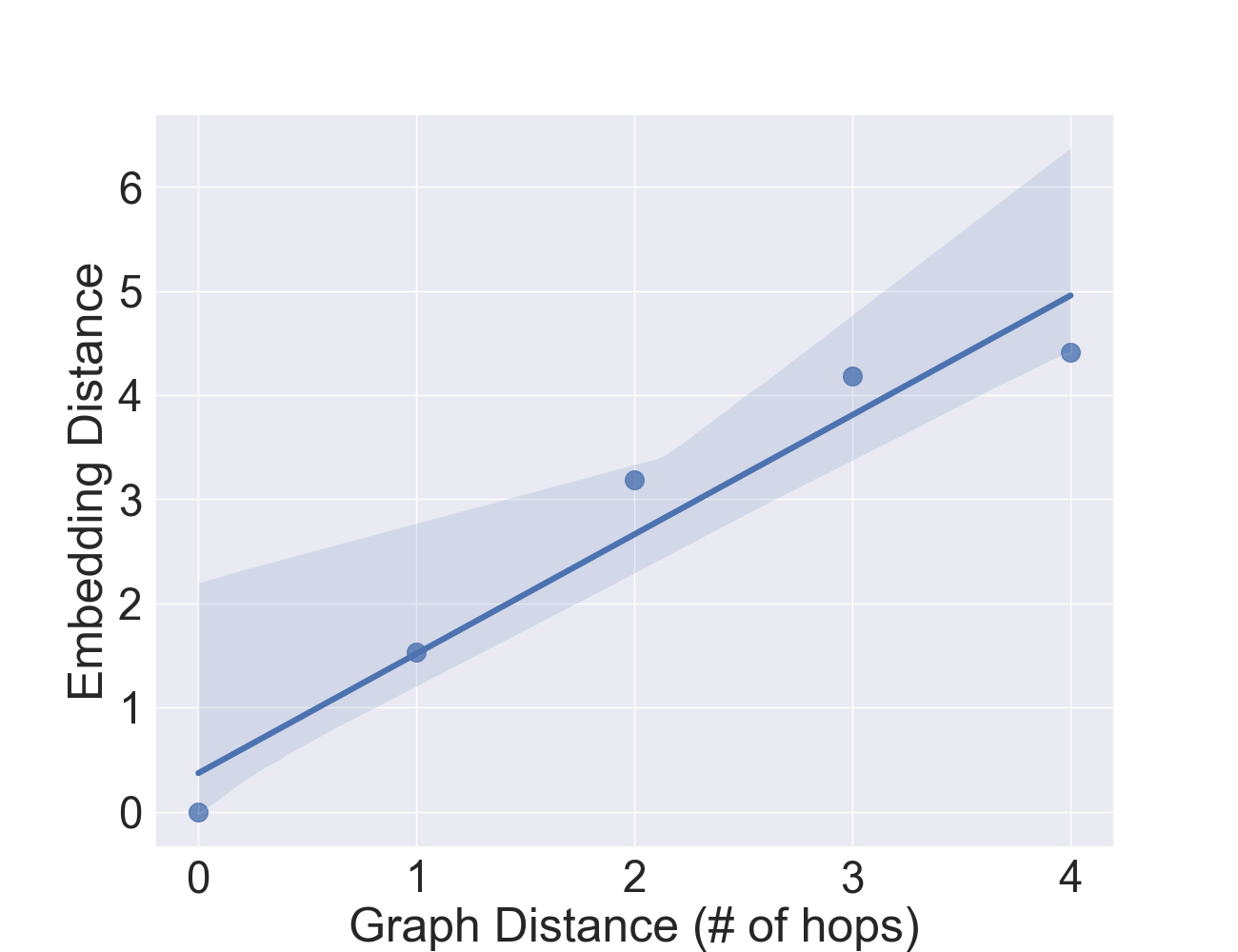}
}
\subfigure[GAT, CoraFull]{
\includegraphics[width=.225\textwidth]{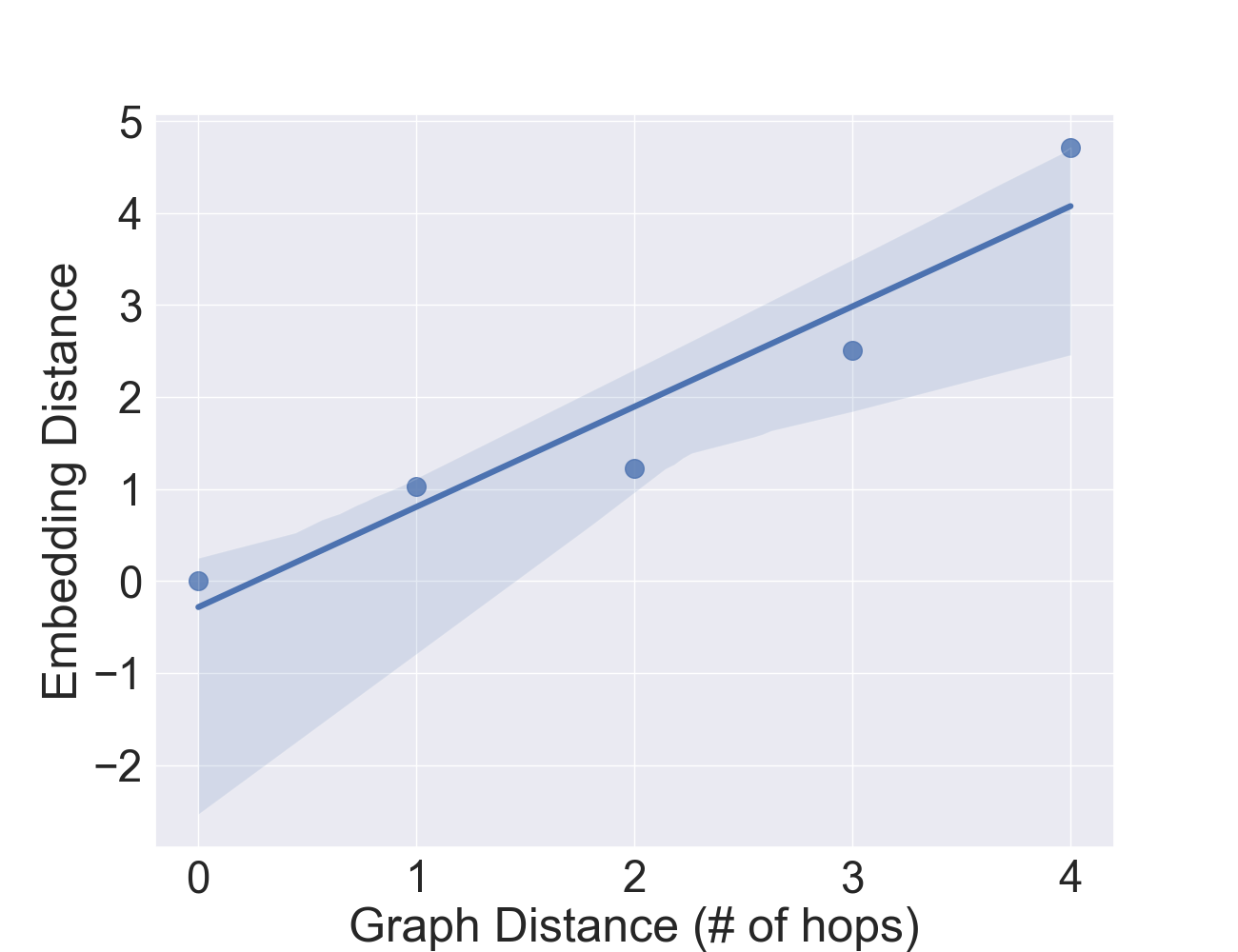}
}
\subfigure[GCNII, CoraFull]{
\includegraphics[width=.225\textwidth]{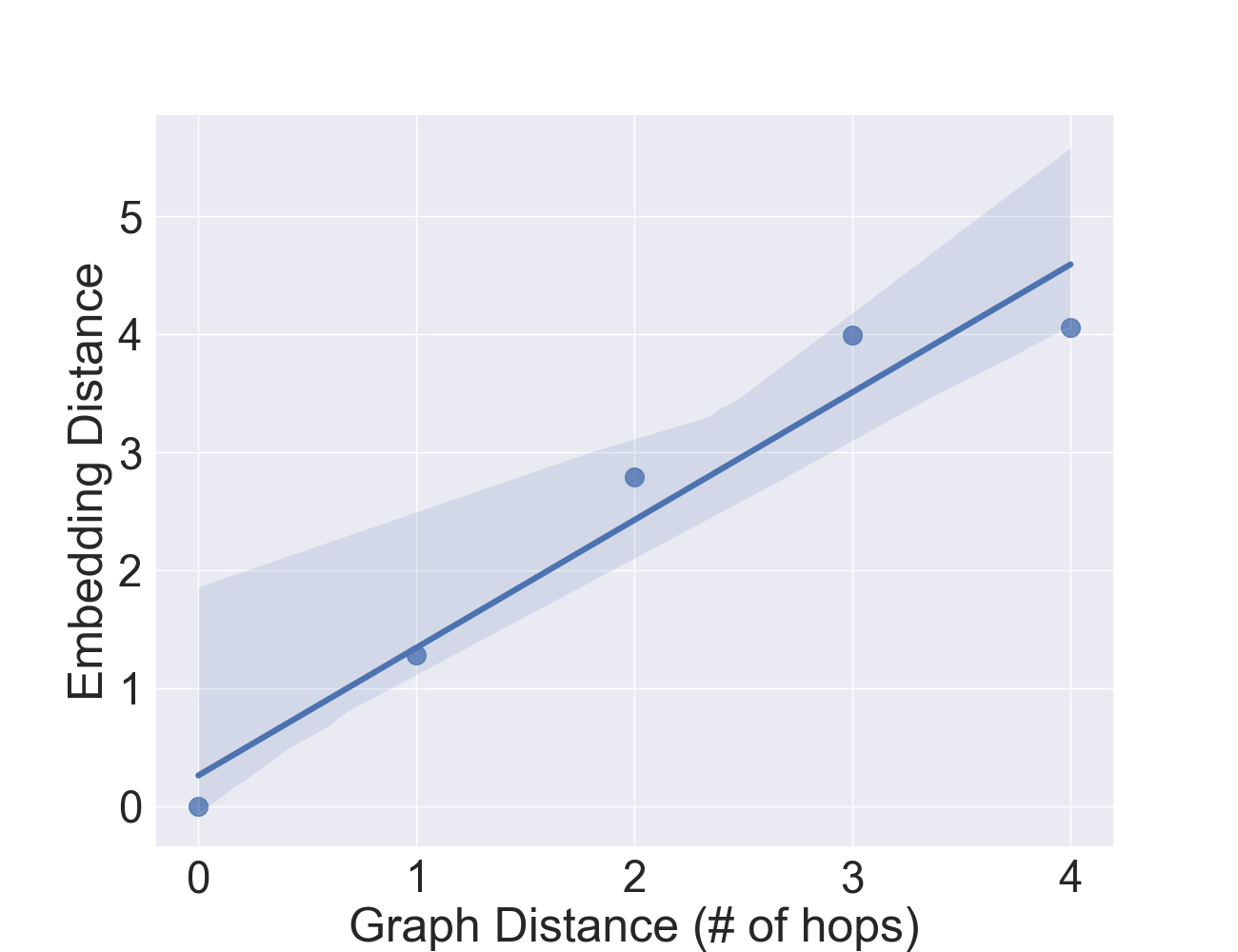}
}
\caption{Graph distance vs.~embedding distance. Additional Results on CoraFull}
\label{fig:group_embedding_citefull}
\end{figure*}

\begin{figure*}
\centering
\subfigure[GCN, OGB-Arxiv]{
\includegraphics[width=.225\textwidth]{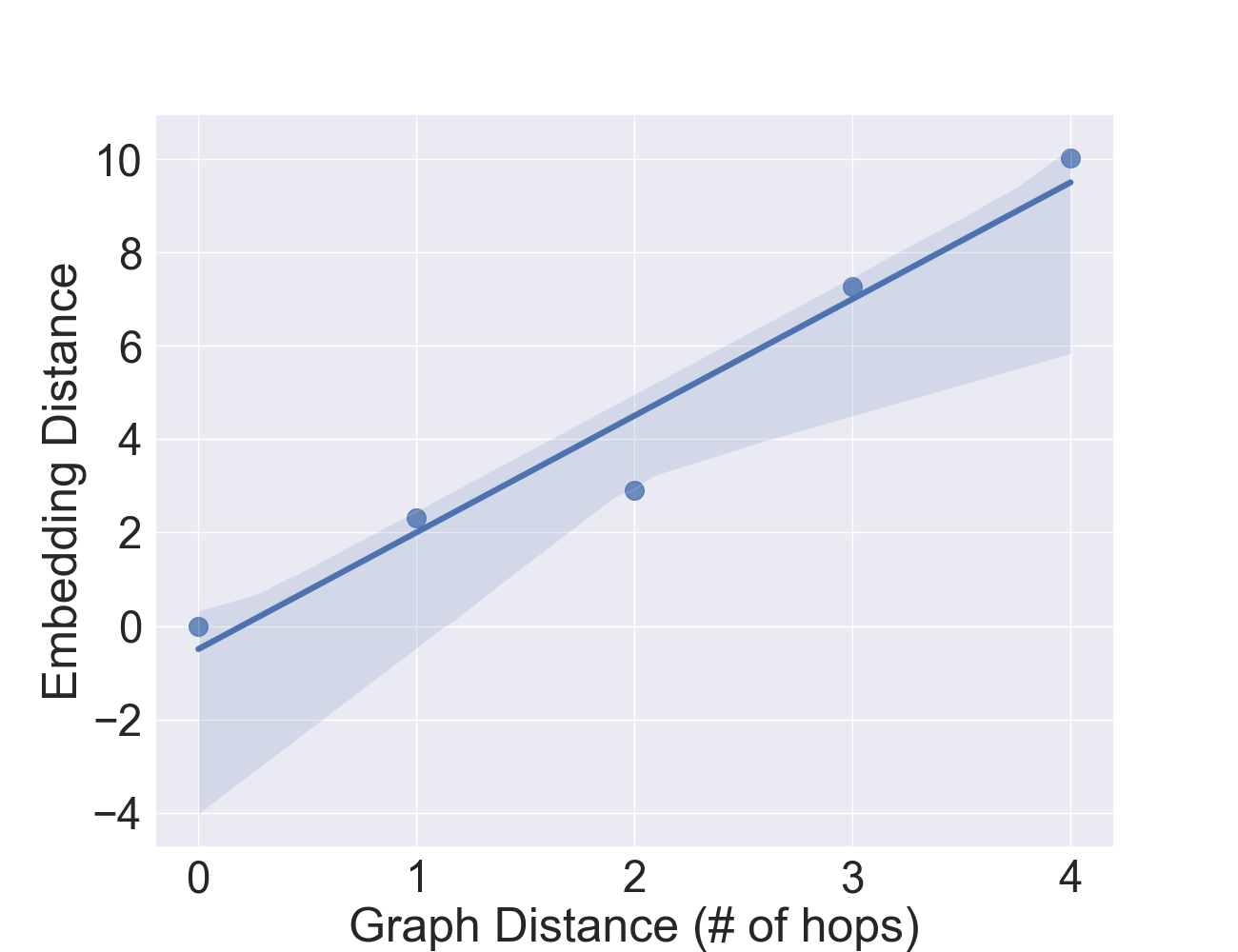}
}
\subfigure[GraphSAGE, OGB-Arxiv]{
\includegraphics[width=.225\textwidth]{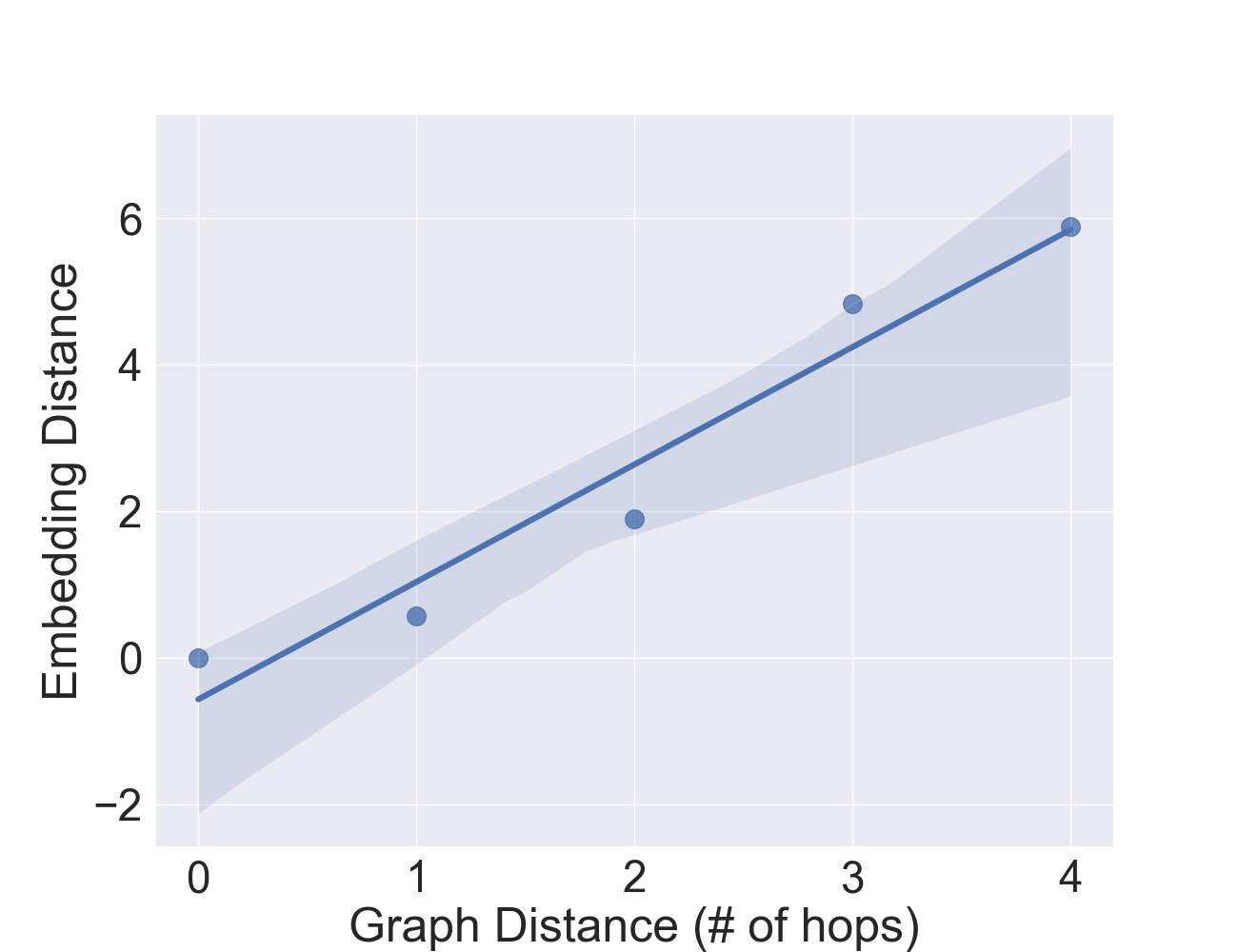}
}
\subfigure[GAT, OGB-Arxiv]{
\includegraphics[width=.225\textwidth]{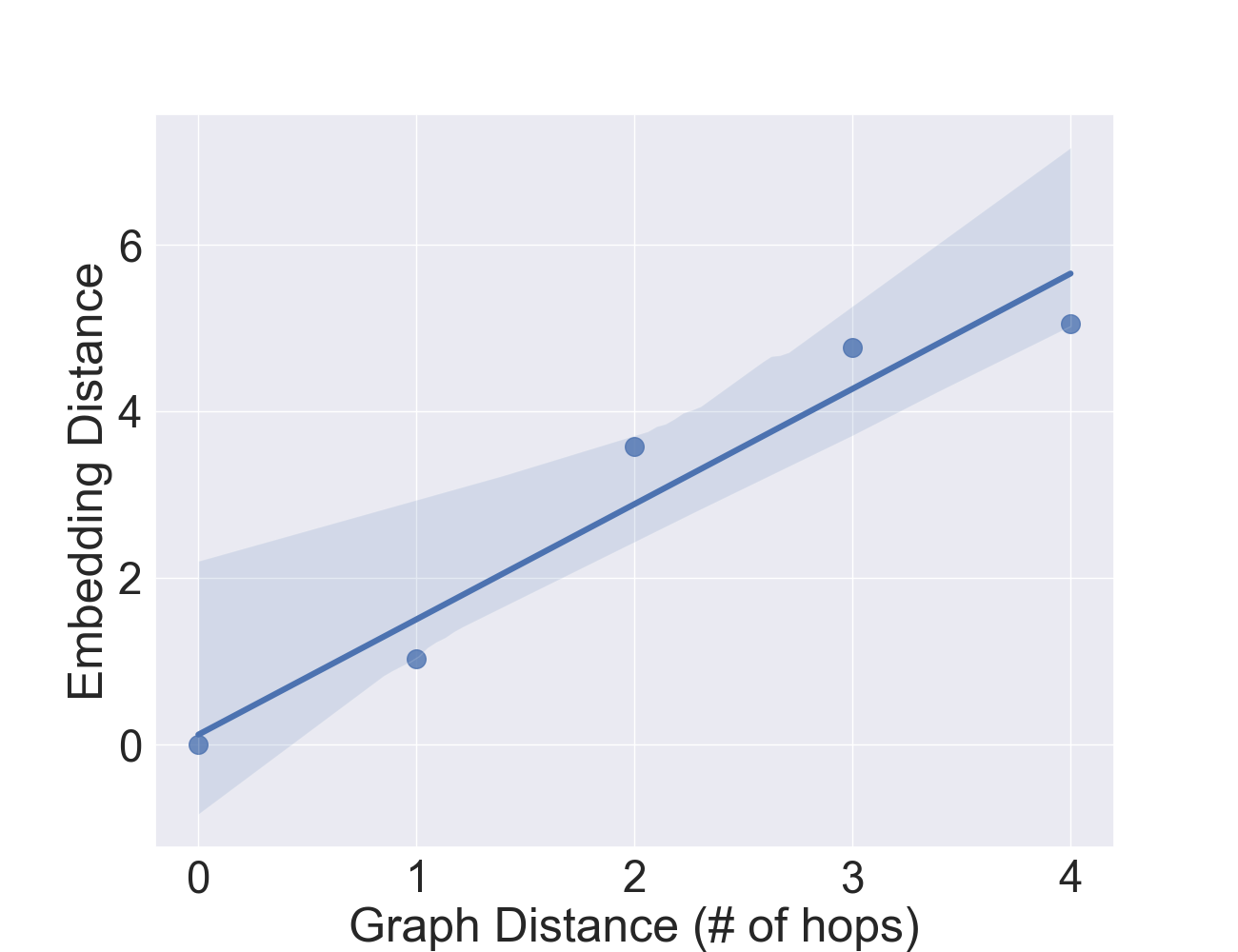}
}
\subfigure[GCNII, OGB-Arxiv]{
\includegraphics[width=.225\textwidth]{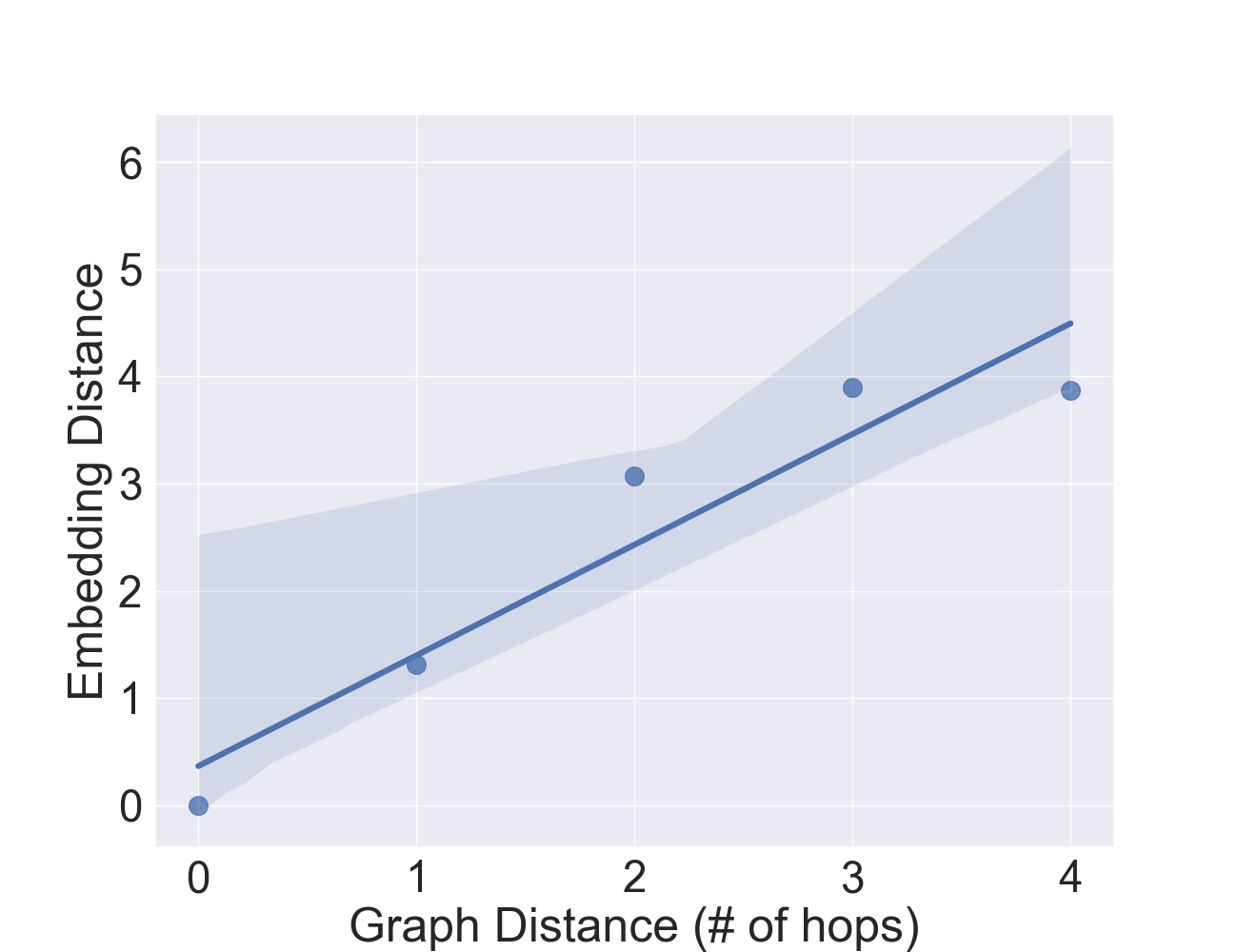}
}
\caption{Graph distance vs.~embedding distance. Additional Results on Ogbn-arxiv}
\label{fig:group_embedding_ogb}
\end{figure*}

We present additional results with Citeseer, CoraFull, and OGBN-Arxiv on that GNNs preserve distances (with a small distortion rate) locally, by evaluating the relation between the graph distances of vertices to the training set and the embedding distances (Euclidean distance) of vertex representations to the representations of the training set, i.e., the relation between $d(v,\mathcal{D})$ and $d(\gnnModel_{\theta_{\mathcal{D}}}(v),\gnnModel_{\theta_{\mathcal{D}}}(\mathcal{D}))$. In this experiment, we follow the same setting as the one in Fig.~\ref{fig:graph_embed_dist} and train the GNN models on the default training set provided in the dataset. We extract vertices within the 5-hop neighbourhood of the training set and group the vertices based on their distances to the training set. We then compute the average embedding distance of each group to the training set in the embedding space. Based on Def.~\ref{def:distortion}, if an embedding has low distortion, we should expect to see linear relation with slop being the scaling factor. As shown in Fig.~\ref{fig:group_embedding_citeseer},Fig.~\ref{fig:group_embedding_citefull}, and Fig.~\ref{fig:group_embedding_ogb} there exists a strong correlation between the graph distance and the embedding distance. The relation is near linear 
and the relative orders of distances are mostly preserved within the first few hops. This implies that the distortion $\alpha$ is indeed small with different GNNs and that the common aggregation mechanism of GNNs can well capture local graph structure information (i.e., graph distances). It also indicates that the premise on the distortion rate, which we used to prove our theoretical results, can be commonly satisfied in practice.

\subsection{Graph Distance and Performance} 
\begin{figure}[!h]
\centering
\includegraphics[width=0.38\textwidth]{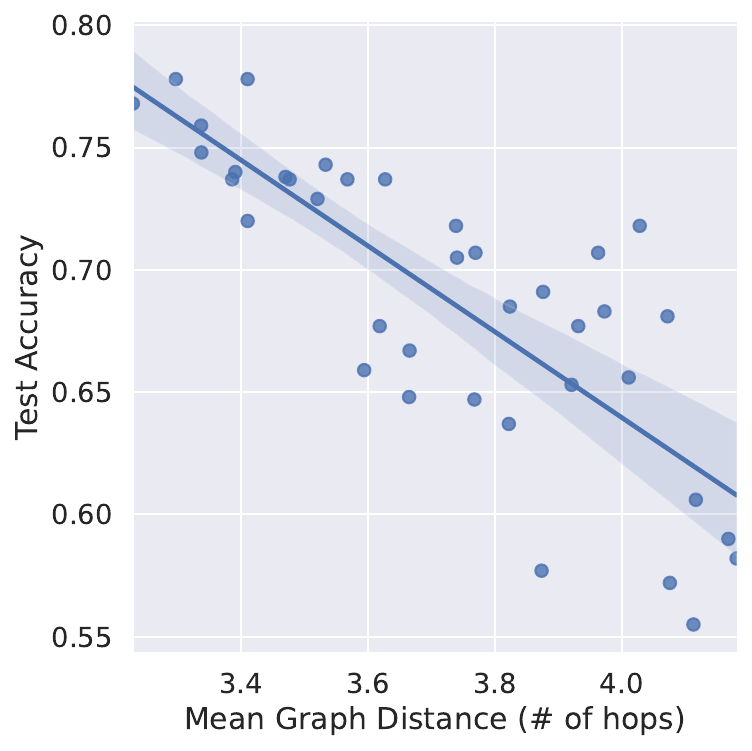}
\hspace{1mm}
\includegraphics[width=0.38\textwidth]{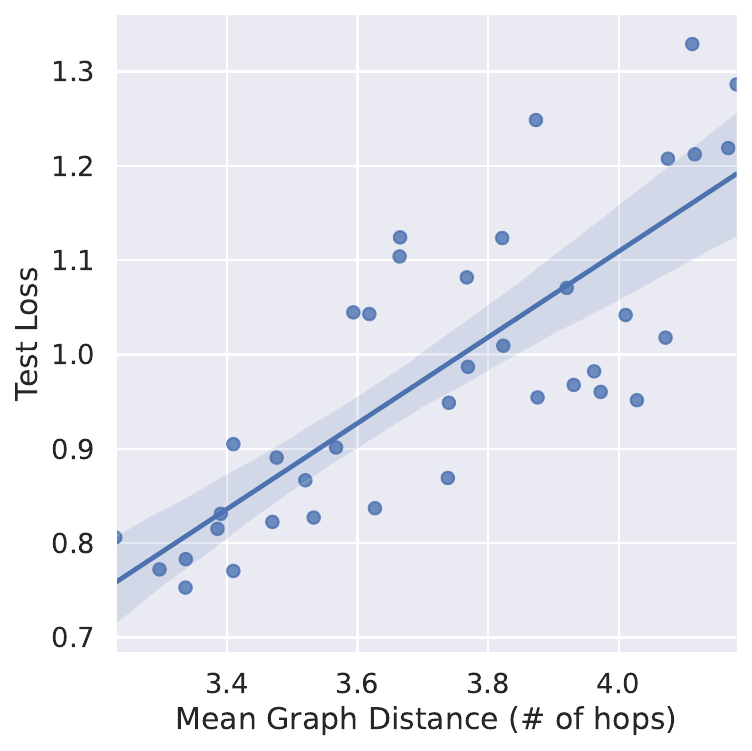}
\caption{{\em Left:} test accuracy vs.~mean graph distance. {\em Right:} test loss vs.~mean graph distance ( to the training set). GCN on Cora.}
\label{fig:main_performance_relation}
\end{figure}

We further explore if the mean distance between different pairs of the training set and test group is related to the prediction performance of GNNs in the test group. In this experiment, we randomly sample 5 vertices from each class in the connected component as the training set $\mathcal{V}_0$. Then we evaluate the GNN model on the rest of the vertices $\mathcal{V}\backslash{V}_0$ in that connected component. The reason for a reduced training set size is to enable a larger variance (among different trials) on the mean distance between the training set and the rest of the vertices. when the training set 
is large, the difference (among different trials) in the mean
distance between the training set and the rest of the vertices to the sampled training set 
is small because of the small-world phenomenon in real-life graph data~\cite{small_world}.  
The mean distance between the training set and the rest of the vertices (i.e., mean graph distance) is computed as $\frac{\sum_{u \in \mathcal{V_0}} d(u,\mathcal{D})}{|\mathcal{D'}|}$. The results are shown in Fig.~\ref{fig:main_performance_relation}. 

As Fig.~\ref{fig:main_performance_relation}, training GNNs on a training set with a smaller mean graph distance to the rest of the vertices, does generalize better, that better performance (test accuracy/loss) is achieved when the distance between the training set and the test group is smaller. This suggests that despite that we use the maximal distance as the definition for our framework, mean graph distance empirically also seems to be a valid indicator for generalization performance. Extending the framework to consider the average distance instead of the maximum distance would be an interesting future work.

\section{Other Potential Applications}\label{appendix:application}
In addition to the initial dataset labelling problem discussed above, there are other potential applications of the proposed framework in other GNNs learning paradigms. Some examples include:

\begin{enumerate}
    \item {\bf Fair k-shot learning}: K-shot learning is a type of machine learning where the model is trained to classify new examples based on a small number of labelled examples in the training set. This type of learning is particularly useful in many situations where the amount of labelled data is limited. In addition to the active learning discussed earlier, k-shot learning can also be applied in choosing a representative experience replay buffer in incremental learning and transfer learning.

    \item {\bf Fairness constraint}: We can apply the derived results to ensure a fair predictive performance of GNNs for specified structural groups. As we have shown in the paper, the key factor of unfair predictive performance towards different structural groups is the distortion between structural distance and embedding distance. We can penalize the parameters with low distortion, which amounts to the following learning objective:
\begin{equation}
    \loss = \loss_{\mathrm{train}} +  \max_{\vertexSet_i,\vertexSet_j \in \vertexSet_1,...,\vertexSet_m} d(\mathrm{GNN}(\vertexSet_i),\mathrm{GNN}(\vertexSet_j)),
\end{equation}
where $\vertexSet_1,...,\vertexSet_m$ are structural subgroups of interest.
    \item {\bf GNNs Profolio}: In this paper, we have focused on analyzing the performance of GNNs with respect to a single structure. However, it is also possible to extend our framework and consider multiple structures, represented by a set $S = \{s_1,...,s_m\}$. The benefits of such an extension are twofold: first, it can provide a more detailed analysis of the GNN's performance; second, the corresponding distortions $\alpha = \{ \alpha_1,...,\alpha_m \}$ can serve as a portfolio of the given GNN, enabling researchers and practitioners to easily determine which GNN is most suitable for a given task.
\end{enumerate}

The proposed framework has promising potential for studying and being applied to the above-mentioned problems. Further research into investigating the actual application results would be an interesting future work.

\end{document}